\documentclass[acmsmall]{acmart}

\usepackage{xspace}
\usepackage{algorithmic}
\usepackage{microtype}
\usepackage{booktabs} 
\usepackage{subfigure}
\usepackage{bm}
\usepackage{amsfonts}
\usepackage{amsmath} 
\usepackage{amstext}
\usepackage{soul}
\setul{0.5ex}{0.3ex}
\setulcolor{blue}
\usepackage{makecell}
\usepackage{nicefrac}       
\usepackage{multirow}

\usepackage{hyperref}       
\usepackage{url}            
\usepackage{booktabs}       
\usepackage{amsfonts}       
\usepackage{nicefrac}       
\usepackage{microtype}      
\usepackage{xcolor}         

\usepackage{enumitem}
\usepackage{nicefrac}   
\usepackage{wrapfig}
\usepackage{setspace}
\usepackage{makecell}
\usepackage{multirow}
\usepackage{calc}
\usepackage{graphicx}
\usepackage{subfigure}
\usepackage{physics}

\usepackage{color, colortbl}
\definecolor{Gray}{gray}{0.9}
\usepackage{ulem}
\usepackage{longtable}
\usepackage{soul}

\usepackage[most]{tcolorbox}
\definecolor{BoxBackground}{RGB}{240, 240, 240} 
\definecolor{BoxFrame}{RGB}{0, 0, 0} 
\definecolor{TitleBackground}{RGB}{0, 0, 0} 
\definecolor{TitleText}{RGB}{255, 255, 255} 
\tcbset{
  academicbox/.style={
    boxsep=5pt,
    left=2pt,
    right=2pt,
    bottom=0.5pt,
    boxrule=0.5pt,
    colback=BoxBackground,
    colframe=BoxFrame,
    colbacktitle=TitleBackground,
    coltitle=TitleText,
    enhanced,
    attach boxed title to top left={yshift=-0.1in,xshift=0.1in},
    boxed title style={boxrule=0pt,colframe=white},
    title={#1},
  }
}

\newtcolorbox{AcademicBox}[1][]{academicbox=#1}
\definecolor{SoftBlue}{RGB}{135, 206, 250}  
\definecolor{SoftOrange}{RGB}{255, 224, 178} 
\definecolor{SoftGreen}{RGB}{144, 238, 144}  
\definecolor{CorrectGreen}{RGB}{76, 175, 80} 
\definecolor{ErrorRed}{RGB}{211, 47, 47} 

\newcommand{\cy}[1]{\textcolor{teal}{{[Rev: #1]}}}

\definecolor{lightcyan}{rgb}{0.88, 1.0, 1.0}
\colorlet{mythmback}{lightcyan!40!white}

\usepackage[customcolors]{hf-tikz} 
\usepackage{soul}
\usepackage{color, xcolor}
\definecolor{lightyellow}{rgb}{1.0, 1.0, 0.6} 

\usepackage{tcolorbox}
\definecolor{lightcyan}{rgb}{0.48, 1.0, 1.0}
\colorlet{mythmback}{gray!40!white}
\newtcolorbox{boxEnv}{
colback=mythmback,coltitle=gray,colframe=mythmback,
center,
width=0.98\linewidth,
boxrule=0.5pt,
left=2pt,right=2pt,
top=2pt,bottom=2pt,
before skip=10pt, after skip=10pt,
fontupper=\footnotesize, 
}

\newtcolorbox{caseBoxEnv}{
colback=white,colframe=black,
center,
width=\linewidth,
boxrule=0.5pt,
left=2pt,right=2pt,
top=2pt,bottom=2pt,
}

\AtBeginDocument{%
  }

\begin{document}

\title{LLMs are Also Effective Embedding Models: \\An In-depth Overview}


\author{Chongyang Tao}
\email{chongyang@buaa.edu.cn}
\orcid{0000-0002-4162-2119}
\affiliation{
  \institution{SKLSDE Lab, Beihang University}
  \streetaddress{37 Xueyuan Road}
  \city{Haidian}
  \state{Beijing}
  \country{China}
  \postcode{100191}
}

\author{Tao Shen}
\email{tao.shen@uts.edu.au}
\affiliation{
  \institution{University of Technology Sydney}
  \streetaddress{15 Broadway, Ultimo NSW 2007 }
  \city{Sydney}
  \country{Australia}
}

\author{Shen Gao}
\email{shengao@uestc.edu.cn}
\affiliation{
  \institution{University of Electronic Science and Technology of China}
  \streetaddress{No.2006, Xiyuan Ave.West Hi-Tech.Zone}
  \city{Chengdu}
  \state{Sichuan}
  \country{China}
  \postcode{611731}
}

\author{Junshuo Zhang}
\email{hanblingz026@gmail.com}
\affiliation{
  \institution{University of Electronic Science and Technology of China}
  \streetaddress{No.2006, Xiyuan Ave.West Hi-Tech.Zone}
  \city{Chengdu}
  \state{Sichuan}
  \country{China}
  \postcode{611731}
}

\author{Zhen Li}
\email{zhenli@pku.edu.cn}
\orcid{0000-0001-9418-8881}
\affiliation{
  \institution{Peking University}
  \streetaddress{No. 5 Yiheyuan Road}
  \city{Haidian}
  \state{Beijing}
  \country{China}
  \postcode{100080}
}

\author{Kai Hua}
\email{kifish.pro@gmail.com}
\affiliation{
  \institution{Peking University}
  \streetaddress{No. 5 Yiheyuan Road}
  \city{Haidian}
  \state{Beijing}
  \country{China}
  \postcode{100080}
}

\author{Wenpen Hu}
\email{wenpeng.hu@pku.edu.cn}
\affiliation{
  \institution{Peking University}
  \streetaddress{No. 5 Yiheyuan Road}
  \city{Haidian}
  \state{Beijing}
  \country{China}
  \postcode{100080}
}

\author{Zhangwei Tao}
\email{tttzw@stu.pku.edu.cn}
\affiliation{
  \institution{Peking University}
  \streetaddress{No. 5 Yiheyuan Road}
  \city{Haidian}
  \state{Beijing}
  \country{China}
  \postcode{100080}
}

\author{Shuai Ma}
\email{chongyang@buaa.edu.cn}
\orcid{0000-0002-4162-2119}
\affiliation{
  \institution{SKLSDE Lab, Beihang University}
  \streetaddress{37 Xueyuan Road}
  \city{Haidian}
  \state{Beijing}
  \country{China}
  \postcode{100191}
}


\begin{abstract}

Large language models (LLMs) have revolutionized natural language processing by achieving state-of-the-art performance across various tasks. Recently, their effectiveness as embedding models has gained attention, marking a paradigm shift from traditional encoder-only models like ELMo and BERT to decoder-only, large-scale LLMs such as GPT, LLaMA, and Mistral. 
This survey provides an in-depth overview of this transition, beginning with foundational techniques before the LLM era, followed by LLM-based embedding models through two main strategies to derive embeddings from LLMs.
1) Direct prompting: We mainly discuss the prompt designs and the underlying rationale for deriving competitive embeddings. 
2) Data-centric tuning: We cover extensive aspects that affect tuning an embedding model, including model architecture, training objectives, data constructions, etc.
Upon the above, we also cover advanced methods for producing embeddings from longer texts, multilingual, code, cross-modal data, as well as reasoning-aware and other domain-specific scenarios.
Furthermore, we discuss factors affecting choices of embedding models, such as performance/efficiency comparisons, dense vs sparse embeddings, pooling strategies, and scaling law. 
Lastly, the survey highlights the limitations and challenges in adapting LLMs for embeddings, including cross-task embedding quality, trade-offs between efficiency and accuracy, low-resource, long-context, data bias, robustness, etc.
This survey serves as a valuable resource for researchers and practitioners by synthesizing current advancements, highlighting key challenges, and offering a comprehensive framework for future work aimed at enhancing the effectiveness and efficiency of LLMs as embedding models. 

\end{abstract}

\ccsdesc[500]{Computing methodologies~Semantic networks}
\ccsdesc[500]{Information systems~Document representation}
\ccsdesc[300]{Information systems~Language models}
\ccsdesc[500]{Information systems~Retrieval tasks and goals}



\keywords{Large Language Models, Information Retrieval, Text Embedding, Prompting, Supervised Finetuning}

\received{10 March 2022}
\received[revised]{4 February 2023}
\received[accepted]{6 February 2023}

    \maketitle

\section{Introduction}

Representation learning is a key concept in deep learning, where models learn to capture meaningful features or patterns from raw data in a compressed, low-dimensional form, known as embeddings~\cite{bengio2013representation,buduma2022fundamentals}. In the context of information retrieval (IR), natural language processing (NLP) and computer vision (CV), representation learning is used to encode a piece of text or images into embedding vectors that capture the semantic meaning and syntactic structure of the input, enabling various downstream tasks such as classification~\cite{muennighoff2022mteb}, retrieval~\cite{muennighoff2022mteb,hao2024toolkengpt}, clustering~\cite{petukhova2024text,petukhova2025text}, anomaly detection~\cite{cao2025tad,gao2025semi}, reward model~\cite{sun2025reusing,sun2025rethinking}, recommendation~\cite{peng2023gpt,liu2024once,zhao2024recommender}, and retrieval-augmented generation (RAG)~\cite{lewis2020retrieval,gao2023retrieval,caspari2024beyond}. 
Embeddings are crucial in modern deep learning literature because they enable efficient representation of high-dimensional data in a compact, dense format~\cite{karpukhin2020dense}. This compression not only reduces storage requirements but also allows for offline computation, which can then be easily used in real-time applications, including retrieval and recommendation systems, with online lightweight operations, e.g., dot-product similarity between embedded vectors.
That is, embeddings preserve essential semantic and syntactic information, making it possible to perform complex operations like similarity comparison or clustering with significantly lower computational overhead.

Attributed to the parallelizable Transformer architecture~\cite{vaswani2017attention} and the availability of high-performance computational resources, representation learning has been moved from shallow-contextualization word2vec~\cite{mikolov2013distributed} to a large-scale pre-training~\cite{devlin2018bert} era in the past years, where models trained on large-scale general corpora could generate generic embeddings that better capture both word- and sequence-level semantics meanings, and could be further enhanced through domain-specific or task-specific fine-tuning. 
Essentially, pre-trained Transformer encoders, including BERT~\cite{devlin2018bert}, RoBERTa~\cite{liu2019roberta} and T5 enocder~\cite{raffel2020exploring}, outperform their RNN-based pioneering works, e.g., CoVe~\cite{mccann2017learned} and ELMo~\cite{peters18deep}, in contextual representation learning, significantly improving performance across various NLP tasks like classification and semantic relatedness.

Nonetheless, pre-training Transformer encoders mainly depend on masked language modeling (MLM), where only a small proportion (e.g., 15\% in BERT) of words or tokens are masked as learning objectives~\cite{devlin2018bert}. 
This deterministic learning process cannot fully perceive the rich contextual dependencies present in unmasked tokens, making these pre-training approaches less efficient and thus quickly reaching performance saturation even scaling model and data size. 
On the other hand, causal language modeling (CLM), learning to predict every next token given its preceding context, is more effective in utilizing both model parameters and training corpora to pre-train generative models~\cite{radford2018improving}, which is proven to have scaling law w.r.t. model performance. 
The resulting LLMs~\cite{brown2020language}, mainly built upon the much deeper Transformer decoder architecture and pre-trained over trillions of tokens, have been proven to have emergent capabilities~\cite{DBLP:journals/tmlr/WeiTBRZBYBZMCHVLDF22} in understanding and reasoning. 
The models, including GPT~\cite{brown2020language,Achiam2023GPT4TR}, LLaMA~\cite{touvron2023llama1,Dubey2024TheL3}, Mistral~\cite{jiang2023mistral} and Qwen~\cite{chu2023qwen,zhang2025qwen3}, achieve remarkable milestones in a broad spectrum of benchmarking tasks with excellent zero-/few-shot capability and state-of-the-art performance, including question-answering, coding, math, reasoning, dialogue generation, etc. 

The success of LLMs has been extended to representation learning, which tunes the LLMs to generate expressively powerful embeddings~\cite{wang2023improving,ma2024fine,lee2024nv,zhuang2024promptreps,nie2024text}. 
Intuitively, the LLMs, with significantly more parameters, larger pretraining corpora, and extended training durations, are expected to outperform the previous BERT-family models in capturing richer semantic representations and contextual nuances. 
This enhanced capacity enables LLMs to generate more accurate and generalizable embeddings, which is expected to improve performance across a wide range of downstream tasks compared to traditional encoder-only models. 

In the context of how to leverage LLMs for representation learning, there are two distinct yet complementary perspectives. 
On the one hand, as LLMs can be viewed as extensions of previous encoder-based models like BERT, several established techniques from the BERT era are still applicable to LLMs. 
For instance, methods such as corpus-aware pre-training~\cite{li2024llama2vec}, multi-task learning~\cite{muennighoff2024generative}, hard negative construction~\cite{wang2023improving}, and distillation from cross-encoder models~\cite{lee2024gecko} are still natural and practical to LLM-based approaches. Some of them have been confirmed in recent studies in terms of effectiveness, which demonstrates the utility of these techniques in fine-tuning LLMs for more robust embeddings. 
However, unique challenges arise when using LLMs for representation learning, particularly in determining how to extract efficient representations from a CLM~\cite{luo2024large}. Unlike BERT's natural use of the \texttt{[CLS]} token to generate embeddings~\cite{devlin2018bert}, extracting useful representations from an LLM trained on CLM objectives~\cite{radford2018improving} is less straightforward.

On the other hand, the powerful expressiveness of LLMs has opened up entirely new paradigms for representation learning, one of which is termed ``Direct Prompt for Embedding''~\cite{jiang2023scaling,zhuang2024promptreps}. 
Thanks to instruction-tuning, many LLMs are endowed with the ability to follow instructions~\cite{ouyang2022training,alpaca,xu2023wizardlm}. This allows practitioners~\cite{jiang2023scaling} to prompt the models to generate a specific category or a topic word, and then utilize its contextualized representations as the final embedding. 
This prompt-based approach presents exciting opportunities but also introduces several open research challenges. For example, the effectiveness and generalization of embeddings generated through direct prompting are still under exploration, with areas like in-context learning~\cite{li2024making} and meta-learning~\cite{lei2024meta} offering potential pathways for improving these embeddings in diverse tasks and domains.

With the growing reliance on LLMs in natural language understanding and information retrieval, this survey aims to provide timely and comprehensive insights into the paradigm shift in embedding methods, offering a deeper understanding of how embedding techniques with LLMs can enhance performance across a wide range of tasks, while also highlighting critical open problems in the field. 
The paper begins with a brief introduction of the foundational techniques that shaped the field before the LLM era. It then introduces two primary approaches: tuning-free embedding methods, which extract meaningful text embeddings from the hidden states of LLMs through direct prompting, without the need for explicit training on embedding-specific tasks; and tuning-based embedding methods, which involve continued supervised fine-tuning, focusing on optimizing model architecture, improving training objectives, and refining training data. 
Next, we summarize advanced techniques developed to handle longer texts, multiple languages, cross-modal data, and codes.
We then compare the performance of various LLM-based embedding methods, considering key aspects in adapting LLMs as embedding models, such as contrasting dense versus sparse embeddings, assessing different pooling strategies, and exploring the implications of scaling laws as LLMs increase in size.
Finally, the survey highlights the limitations and emerging challenges involved in adapting LLMs to be more effective as embedding models.

\section{Background}

Over the past decade, the paradigms for representation learning have shifted multiple times because of the rapid advancements in neural architectures, and the availability of large-scale datasets and computational resources: 
the first being shallow contextualization (e.g., word2vec and specific tuning), the second marked by BERT's pre-training methods, and now, the third transition towards LLMs as embedding models. 
This section provides a brief overview of foundational knowledge and techniques that shaped the field before the LLM era.

\subsection{Shallow Contextualization}

Initially, word-level representations were learned using shallow models like word2vec~\cite{mikolov2013distributed,mikolov2013efficient}, GloVe~\cite{pennington2014glove}, and FastText~\cite{bojanowski2017enriching}, which employed techniques such as skip-gram or continuous bag-of-words (CBoW) to capture context, with sequence-level representations typically obtained through a weighted sum of word vectors, where the weights were calculated using heuristic methods like TF-IDF~\cite{das2018improved}.
However, these methods' shallow architecture limits their ability to fully leverage contextual information and struggle with capturing polysemy and ambiguity, leading to unsatisfactory performance in a wide range of NLP and IR tasks. 
On the other hand, some follow-up works constructed unsupervised training objectives at sequence (e.g., sentence) level, such as Skip-Thought~\cite{kiros2015skip}. 
Later, contextualized models like CoVe~\cite{mccann2017learned} and ELMo~\cite{peters18deep} addressed these limitations by using bi-directional LSTMs to capture richer contextual information and generate word representations that adapt to the surrounding context. ELMo, in particular, significantly improved tasks requiring nuanced semantic understanding due to its contextual flexibility.

\subsection{Prominent Techniques in BERT Era}

The advent of BERT~\cite{devlin2018bert} and its successors, such as RoBERTa~\cite{liu2019roberta} and T5-encoder~\cite{raffel2020exploring}, marked a significant leap in representation learning, leveraging large-scale pre-training to capture deep contextualized embeddings that could be adapted to different domains and tasks. In the following, we will introduce several prominent techniques to improve embedding quality, which have the potential to benefit LLMs as embedding models.

\vspace{1mm}
\textbf{Corpus-aware pre-training.} These pre-trained models are all trained on general text corpora, but their performance on specific target domains is often suboptimal. A straightforward approach to improve domain-specific performance is to do continual pre-training over the corpus in the target domain~\cite{izacard2021unsupervised,gao2021condenser}. The training objectives can be quite flexible, ranging from general tasks like masked language modeling or next sentence prediction to more specialized methods that generate pseudo-contrastive learning examples through heuristics. Furthermore, research has shown that a novel bottleneck-enforced pre-training approach can be effective~\cite{gao2021unsupervised,wang2022simlm,xiao2022retromae,shen2022lexmae}, where the original encoder structure is retained but a weaker decoder is introduced. This decoder reconstructs the original input based on a bottleneck, i.e., a single-vector embedding from the encoder, which forces semantic knowledge to be retained in the embedding.

\vspace{1mm}
\textbf{Hard negative mining.}
Negative samples are essential in contrastive learning, as they are examples that, compared to the positive sample, are far from the anchor according to a specific metric. In contrast to random negative samples, learning with hard negatives has proven to be highly effective in representation learning for embeddings~\cite{xiong2020approximate,wang2022simlm,shen2023unifier,tao2024adam}. Intuitively, hard negatives, which are closer to the anchor but belong to different classes, force the model to make more refined distinctions, thus leading to more robust and accurate embeddings. Therefore, how to construct or sample hard negatives has remained a popular research topic, with methods like using a BM25 retriever or the strongest available retriever to select challenging negatives from large-scale collections.

\vspace{1mm}
\textbf{Supervision from re-ranker.}
In model training, knowledge distillation from a teacher (stronger or larger) model to a student model (weaker or smaller) has been proven effective in various scenarios and tasks of deep learning~\cite{xu2024survey}. 
In the field of embedding, one unique opportunity is to distill knowledge from a cross-encoder-based re-ranker to a bi-encoder-based encoding/embedding model by applying a KL divergence loss between the score distribution over several candidates against their anchor~\cite{qu2020rocketqa}. 
Here, a cross-encoder~\cite{zhou2022towards} directly computes interaction between every pair of input sequences, resulting in richer contextual embeddings but with higher computational cost, while a bi-encoder~\cite{karpukhin2020dense} independently encodes two sequences and computes their similarity in the embedding space, offering a more efficient yet slightly less expressive representation. 
The supervision from the cross-encoder allows the bi-encoder to learn more nuanced distinctions and improve its embedding quality. 
Furthermore, studies have shown that mutual supervision or distillation between cross- and bi-encoders can simultaneously lead to boosted re-ranking performance for the cross-encoder and enhanced encoding for the bi-encoder, creating a mutual learning framework that benefits both models~\cite{ren2021rocketqav2,feng2022reciprocal,cai2022hyper}.

\vspace{1mm}
\textbf{Multi-task learning.} 
Multi-task learning is a machine learning approach where a model is trained to perform multiple tasks simultaneously, leveraging shared information across tasks to improve its overall learning efficiency and generalization~\cite{liu2019multi,raffel2020exploring,sanh2021multitask}. This approach is effective because it allows the model to capture commonalities between tasks while also learning task-specific nuances, leading to more robust performance across diverse scenarios. In the field of embedding models, multi-task learning has shown great potential for improving generalization and robustness by enabling models to learn representations that are not only effective for a single task but adaptable across various tasks~\cite{maillard2021multi,cai2022hyper}. This adaptability helps the model capture richer and more diverse semantic features, making the embeddings more versatile for downstream applications like classification, retrieval, and clustering. 
Going beyond and following the inspiration of prompt-tuning~\cite{lester2021power,liu2021p}, some pioneering works~\cite{su2023one,xiao2023bge,zhang2023retrieve} propose to unify a broad spectrum of retrieval tasks by augmenting a piece of text with an instruction/explanation of the corresponding task or domain, which is still the most prevalent paradigm up to now.

In the remaining sections, starting with a task formulation, we will explore extensions and variants of prior techniques in the LLM era, as well as advancements and challenges in utilizing LLMs as effective embedding models. 
In general, we will delve into both direct prompting and fine-tuning approaches, and provide insights into their performance across various downstream tasks.

\begin{figure*}[ht!]
\centering
\includegraphics[width=0.95\linewidth]{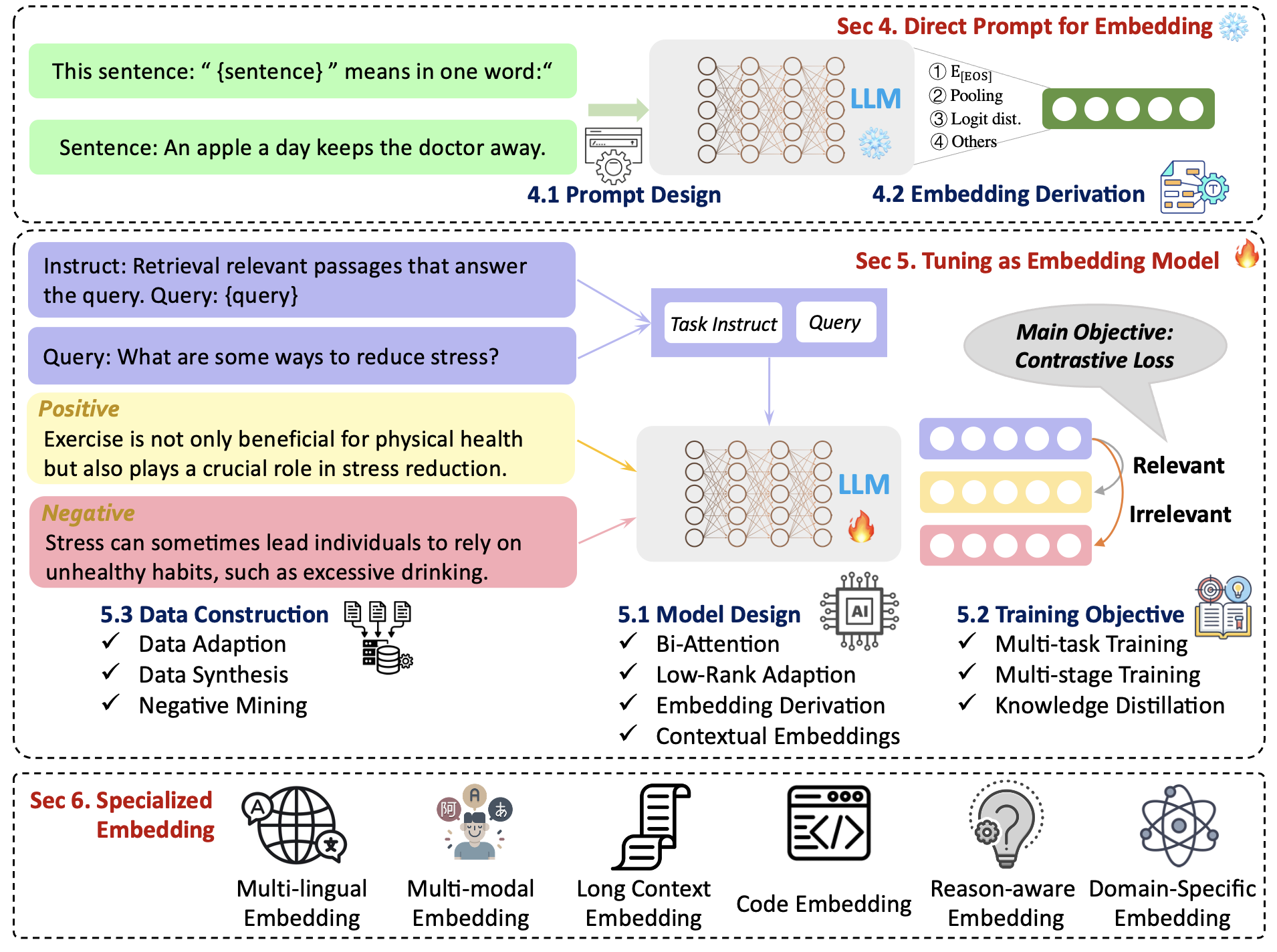}
\caption{An overview of our survey on deploying LLMs as embedding models, covering both \textcircled{1} direct prompting as embedding models, \textcircled{2} fine-tuning as embedding models and \textcircled{3} specialized embedding mdels.}
\label{fig:llm4emb}
\end{figure*}

\section{Problem Formalization and Survey Overview}
\label{sec:future}

Let \( \mathbf{x} = \{x_1, x_2, \dots, x_n\} \) be an input sequence of tokens where \( x_i \) represents the \( i \)-th token in the sequence. The decoder-only LLM \( \mathcal{M} \) processes this sequence to produce a contextualized representation for each token, denoted as
\begin{align*}
\{\mathbf{h}_1, \mathbf{h}_2, \dots, \mathbf{h}_n\} = \mathcal{M}(\mathbf{x}, \theta), \\
\mathbf{v} = f_{\text{agg}}(\{\mathbf{h}_1, \dots, \mathbf{h}_n\}),
\end{align*}
where \( \mathbf{h}_i \) is the output hidden state or logit distribution for token \( x_i \),
a fixed-length embedding vector $ \mathbf{v}$  for the entire sequence can be obtained by the aggregation function $f_{\text{agg}}$, such as special-token selection (e.g., [EOS]), mean pooling, or attention-based weighting.

This formulation provides a general framework for using decoder-only LLMs as embedding models. However, practical instantiations of this process vary widely depending on how the model is used, adapted, or trained. As illustrated in Figure~\ref{fig:llm4emb}, recent research has proposed three major paradigms for generating embeddings from LLMs: {direct prompting}, {fine-tuning}, and {specialization}. 
First, we explore how LLMs can be directly prompted to produce embeddings, with a focus on prompt design and embedding derivation strategies that require no parameter updates. Second, we examine tuning-based methods that convert LLMs into dedicated embedding models through architectural adaptations (e.g., bi-attention, low-rank adaptation), refined embedding extraction mechanisms, and training objectives such as multi-task learning, multi-stage optimization, and knowledge distillation. These approaches are supported by carefully constructed datasets, leveraging both adaptation of existing data and synthesis via LLMs, often with hard negative mining to improve contrastive learning. Finally, we discuss specialized embeddings developed for multilingual, multimodal, long-context, code, reason-aware, and domain-specific scenarios, which address real-world needs and highlight the flexibility of embedding models in capturing complex semantics beyond general-purpose tasks.

In addition to these paradigms and techniques, this survey examines critical considerations in adapting LLMs as embedding models, including the comparison of dense and sparse embeddings, evaluation of various aggregation strategies, and analysis of scaling law implications. Furthermore, we also highlight open challenges and future directions, shedding light on key limitations and potential research opportunities in LLM-based embedding techniques.

\section{Tuning-Free Embedding Models}

\begin{figure}[t!]
    \begin{AcademicBox}
    \small
    \vspace{-2mm}
    \textbf{\emph{PromptEOL}}~\cite{jiang2023scaling}\ \\
    \texttt{This sentence : "[X]" means in one word:\color{red}{"}}\\
    
    \textbf{\emph{PromptSTH}}~\cite{zhang2024simple}\\
    \texttt{This sentence : "[X]" means \color{red}{something}}\\
    
    \textbf{\emph{PromptSUM}}~\cite{zhang2024simple}\\
    \texttt{This sentence : "[X]" can be summarized \color{red}{as}}\\
    
    \vspace{2pt}  \hrule \vspace{6pt}

    \textbf{\emph{Pretended Chain of Thought}}~\cite{zhang2024simple} \\
    \texttt{After thinking step by step , this sentence : “[X]” means in one word:“} \\
    
    \textbf{\emph{Knowledge Enhancement}}~\cite{zhang2024simple} \\
    \texttt{The essence of a sentence is often captured by its main subjects and actions,
    while descriptive terms provide additional but less central details.    With this in mind , this sentence : “[X]” means in one word:“}
    

    \end{AcademicBox}
    \caption{Prompts used in various tuning-free methods.}
    \label{fig:PromptEOL}
\end{figure}

Previous masked language models, like BERT and RoBERTa, employ a mask prediction task to capture contextual information for specific tokens. Building on this factual, PromptBERT~\cite{jiang2022promptbert} frames the text embedding extraction as a similar task, utilizing the following template for prompting: \colorbox{gray!30}{\texttt{This sentence : "[X]" means [MASK] .}}
In this template, [X] and [MASK] represent placeholders for the input sequence and the mask token, respectively. The hidden vector of the [MASK] token from the final layer is directly used as the representation of the text sequence.

Recently, researchers have started to explore the potential of extracting meaningful text embeddings directly from the hidden states of LLMs (e.g., OPT or LLAMA) without requiring explicit training on embedding-specific tasks. These embeddings have been applied in various tasks, including clustering~\cite{petukhova2024text}, recommendation~\cite{peng2023gpt,liu2024once,li2023exploring}, and retrieval~\cite{lei2024meta,zhuang2024promptreps}. 
These methods typically involve using fill-in-the-blank prompts. Inspired by PromptBERT, \cite{jiang2023scaling} introduced PromptEOL, which enhances the prompt-based method in BERT by incorporating an "explicit one word limitation" (EOL) to extract text representations for LLMs. As shown in the first line of Figure~\ref{fig:PromptEOL},
PromptEOL incorporates ": ``" at the end of the template to prevent the model from generating punctuation in the next token. Additionally, it considers an in-context learning framework to automatically create and search for demonstration sets to improve embeddings in LLMs.  This method improves performance across all OPT models, allowing them to match or even outperform BERT in prompt-based embedding tasks.

More recently, \cite{zhang2024simple} reveals that EOL does not consistently yield optimal performance when fine-tuning generative models. They introduced two templates, PromptSTH and PromptSUM, which intentionally omit the “in one word” constraint, along with two powerful prompt engineering methods, Pretended CoT and Knowledge Enhancement, to enable the model to analyze diverse semantic aspects, as shown in Figure~\ref{fig:PromptEOL}.
To effectively capture multiple representations of sentences from distinct perspectives, \cite{lei2024meta} introduced MetaEOL, which utilizes a diverse set of meta-task prompts, including text classification, sentiment analysis, paraphrase identification, and information extraction, to generate embeddings. 
\cite{thirukovalluru2024geneol} further present GenEOL, which leverages LLMs to generate diverse transformations of a sentence that retain its original meaning, and then aggregates the embeddings of these transformations to enhance the final sentence embedding.
\cite{zhuang2024promptreps} directly prompt LLMs to generate both dense embedding representations and sparse bag-of-words representations for document retrieval. 
\cite{springer2024repetition} propose echo embeddings that repeat the input twice in context and extract embeddings from the second occurrence. 
\cite{li2024your} introduced MOE Embedding (MOEE), which combines the routing weights and hidden states of Mixture-of-Experts LLMs to form a powerful zero-shot embedding model that surpasses traditional approaches without requiring any additional fine-tuning.

\section{Tuning-based Embedding Models}
While direct prompting of LLMs can yield useful embeddings, the true potential of these models is unlocked through a more refined approach: tuning them with existing or synthetic paired text data. 
This process typically employs contrastive learning to enhance the models' ability to distinguish between semantically similar and dissimilar text pairs, resulting in more accurate and meaningful embeddings. For example, in the early stage of LLMs outbreak, \cite{muennighoff2022sgpt} initialized the embedding models with pre-trained GPT-3 models and applied continued contrastive training. The hidden state from the last layer corresponding to the special token [EOS] at the end of the sequence is taken as the embedding of the input sequence. Instructor~\cite{su2023one} is the first work to explore a unified approach for generating text embeddings using task instructions based on GTR~\cite{ni2021large}. It annotates instructions for 330 diverse tasks and trains the model on this multitask mixture with a contrastive loss.

With the continuous progress of open-source LLMs, many researchers have attempted to use these models to build better text representation models. These advances primarily focus on  \emph{optimizing model architecture}~\cite{lee2024gecko,muennighoff2024generative},  \emph{improving training objectives}~\cite{behnamghader2024llm2vec,li2023towards,lee2024nv,lee2024gecko}, and further \emph{enhancing training data}~\cite{wang2023improving,kim2024linq,jeronymo2023inparsv2}. In this section, we will delve into these three foundational techniques that underpin the fine-tuning of LLMs as powerful representation models. Table~\ref{tab:config} summarize  summarizes the configurations of representative models in terms of these three aspects.

\subsection{Model Architecture}

\begin{figure*}[ht!]
\centering
\includegraphics[width=\linewidth]{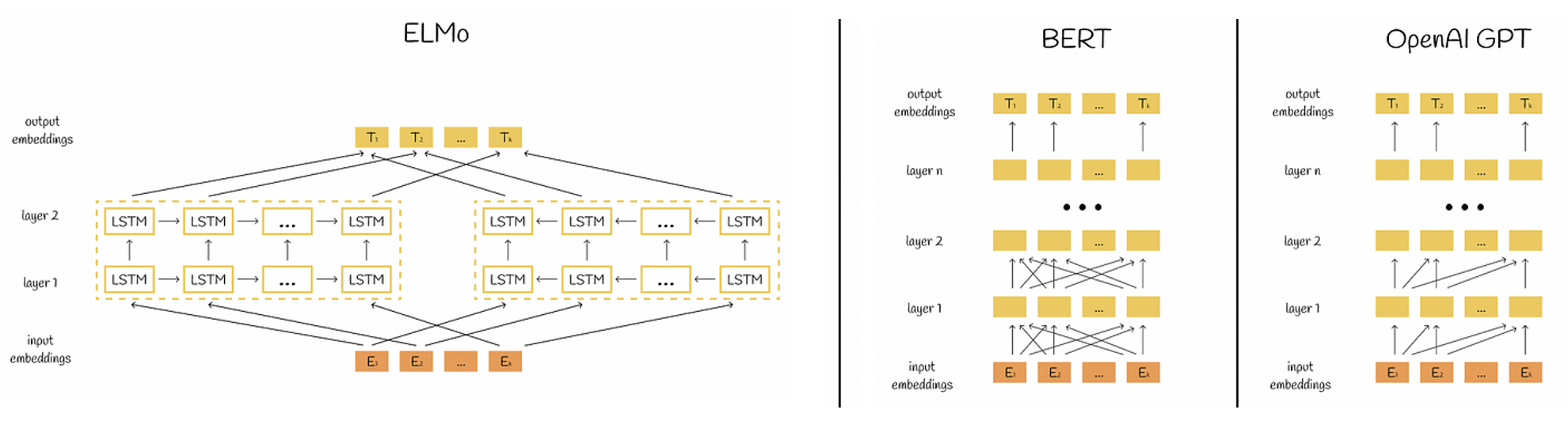}
\caption{An illustration of the evolution of text embeddings from ELMo and BERT to GPT-style models, from \cite{zaib2020short}.}
\label{fig:bert_gpt}
\end{figure*}

The development of language models has seen significant changes in neural architecture design. As illustrated in Figure~\ref{fig:bert_gpt}, early models like ELMo~\cite{peters18deep} and BERT~\cite{devlin2018bert}  primarily utilized encoder-only architectures, which focus on bidirectional context, enabling the model to consider both left and right context during training. 
In contrast, recent LLMs have adopted decoder-only architectures~\cite{openai2023gpt4,touvron2023llama,jiang2023mistral}, which are driven by the need for models to excel at generative tasks, such as text completion and creative writing. While encoder-only models provide strong contextual embeddings, decoder-only models have demonstrated superior performance in generative tasks due to their autoregressive nature.

\vspace{1mm}
\textbf{Embedding with Bi-directional Contextualization.}
Current generative LLMs predominantly use mono-directional attention, focusing on a unidirectional flow of information. This approach simplifies the model architecture and aligns well with autoregressive tasks where future tokens are predicted based on past context. However, the lack of bidirectional attention can limit the model's ability to fully capture dependencies within the entire sequence. To address this, some models like Gecko~\cite{lee2024gecko} and LLM2vec~\cite{behnamghader2024llm2vec} propose incorporating bidirectional attention mechanisms within existing LLMs, enabling the model to consider both past and future tokens simultaneously. This enhancement aims to improve the quality of sequence embeddings by leveraging a more comprehensive understanding of the input text.
In addition, GritLM~\cite{muennighoff2024generative} unifies embedding tasks and generative tasks into a single model with bidirectional attention through generative representational instruction tuning.
Recently, however, BGE-ICL~\cite{li2024making} argue that enabling bidirectional attention during embedding fine-tuning misalign with the model's original pre-training setup, which could compromise its effectiveness in generation tasks.

\begin{table*}[t!]
\centering
\caption{Detail of representative models that tune LLMs as embedding models. ``sup.CL" and ``unsup.CL" means supervised contrastive loss and unsupervised contrastive loss respectively. ``Bi-Att" indicates whether the model enables bi-directional attention in LLMs, while ``LoRA" specifies whether the model was trained using the LoRA technique. 
For the manner of embedding derivation, ``MP" is mean pooling operation and ``P-MP" means Position-weighted mean pooling, ``LAT" means Latent Attention Layer.}

\setlength{\tabcolsep}{1.8pt}
\resizebox{1.0\linewidth}{!}{%
\begin{tabular}{lccc|ccccc}
\toprule
Model & Dim. & \# Params. & Base & Fusion & Bi-Att & LoRA & Training Data &  Training Obj. (Neg) \\ \midrule
		
SGPT~\cite{muennighoff2022sgpt} & 4,096 & 5.8B & GPT&  P-MP & $\times$ & $\times$ & SNLI/MNLI &  sup.CL \\
Instructor-XL~\cite{su2023one} & 768 & 1.5B & GTR & MP & $\times$ & $\times$ & MEDI & sup.CL \\
 \midrule
 GTE-Qwen2-7B~\cite{li2023towards} &3,584& 7B &  Qwen2  & MP & $\checkmark$ & $\times$ &  Unsup/Sup Pair & unsup.CL \& sup.CL\\ 

E5-mistral-7b~\cite{wang2023improving} & 4,096 & 7B & Mistral & [EOS] & $\times$ & $\checkmark$ & E5(Public\&Synthetic) &  sup.CL \\ 

{GritLM-7B}~\cite{muennighoff2024generative} & 4,096 & 7B & Mistral & MP & $\checkmark$ & $\times$ & E5S(E5\&S2ORC)/Tülu 2 & sup.CL \& NTP \\

Echo-mistral-7b~\cite{springer2024repetition} & 4,096 & 7B & Mistral & [EOS] & $\times$ & $\checkmark$ & E5 & sup.CL\\

LLM2Vec~\cite{behnamghader2024llm2vec} & 4,096 & 8B & Llama3 & MP & $\checkmark$ & $\checkmark$ & Wikipedia & NTP \& unsup.CL \\  
 
SFR-Embedding & 4,096 & 7B & E5 &  [EOS] & $\times$ & $\times$ & Specially Curated Dataset & sup.CL\\ 

NV-Embed~\cite{lee2024nv} & 4,096& 7B & Mistral & LAT & $\checkmark$ & $\checkmark$ & Public Sup Datasets & Two-stage sup.CL\\
Linq-Embed-Mistral~\cite{kim2024linq} & 4,096 & 7B &  E5 &  [EOS] & $\times$ & $\checkmark$ & E5S\&Refined Synthetic & sup.CL \\
{Gecko}~\cite{lee2024gecko} & 256 & 1.2B & Transformer & MP & $\checkmark$ & $\times$ &  FRet\&Public Datasets  &  sup.CL \\ 
NV-Retriever~\cite{moreira2024nv} & 4096 &7B & Mistral & MP & \checkmark & \checkmark & Public Sup Datasets & Sup CL \\

BGE-ICL~\cite{li2024making} & 4,096 & 7B & Mistral & [EOS] & $\times$ & $\checkmark$ & Public Sup Datasets & sup.CL \& KD \\

Gemini-Embedding~\cite{lee2025gemini} & 3072 & UNK & Gemini & MP & $\times$ & $\times$&  Pub \& Syn & sup.CL \\
Qwen3-Embedding~\cite{zhang2025qwen3} & 4096 & 4B/8B & Qwen3  & [EOS]  & $\times$ & $\times$ & Pub \& Syn & sup.CL \\

    \bottomrule
    \end{tabular}
}
\label{tab:config}
\end{table*}

\vspace{1mm}
\textbf{Low-Rank Adaption.} Traditional fine-tuning methods often require extensive computational resources and large amounts of labeled data. As a result, there is a growing interest in parameter-efficient tuning approaches (e.g., LoRA) that enable the adaptation of LLMs for embedding tasks while minimizing resource consumption~\cite{wang2023improving,kim2024linq,li2024making}. These parameter-efficient tuning methods offer promising solutions by reducing the number of trainable parameters while maintaining comparable model performance or even better generalization performance~\cite{wang2023improving}.

\vspace{1mm}
\textbf{Embedding Derivation. }
Obtaining sequence embeddings from these architectures involves different techniques. In BERT, the output states of [CLS] token is commonly used as a representation of the entire input sequence, providing a summary embedding. For LLMs, the [EOS] token, which signifies the end of a sequence, serves a similar purpose. Additionally, mean pooling of the last hidden layer can be employed to aggregate the contextual information across all tokens in the sequence~\cite{su2023one,springer2024repetition,behnamghader2024llm2vec,lee2024gecko}.Recent studies have explored more advanced methods to obtain sequence representations, such as using latent attention layer~\cite{lee2024nv} that adaptively combine token embeddings to enhance the final sequence representation, or sparse representations based on lexicon-importance distribution~\cite{chen2024bge,zhuang2024promptreps,doshi2024mistral,lupart2024disco}.



\begin{figure*}[ht!]
\centering
\includegraphics[width=0.6\linewidth]{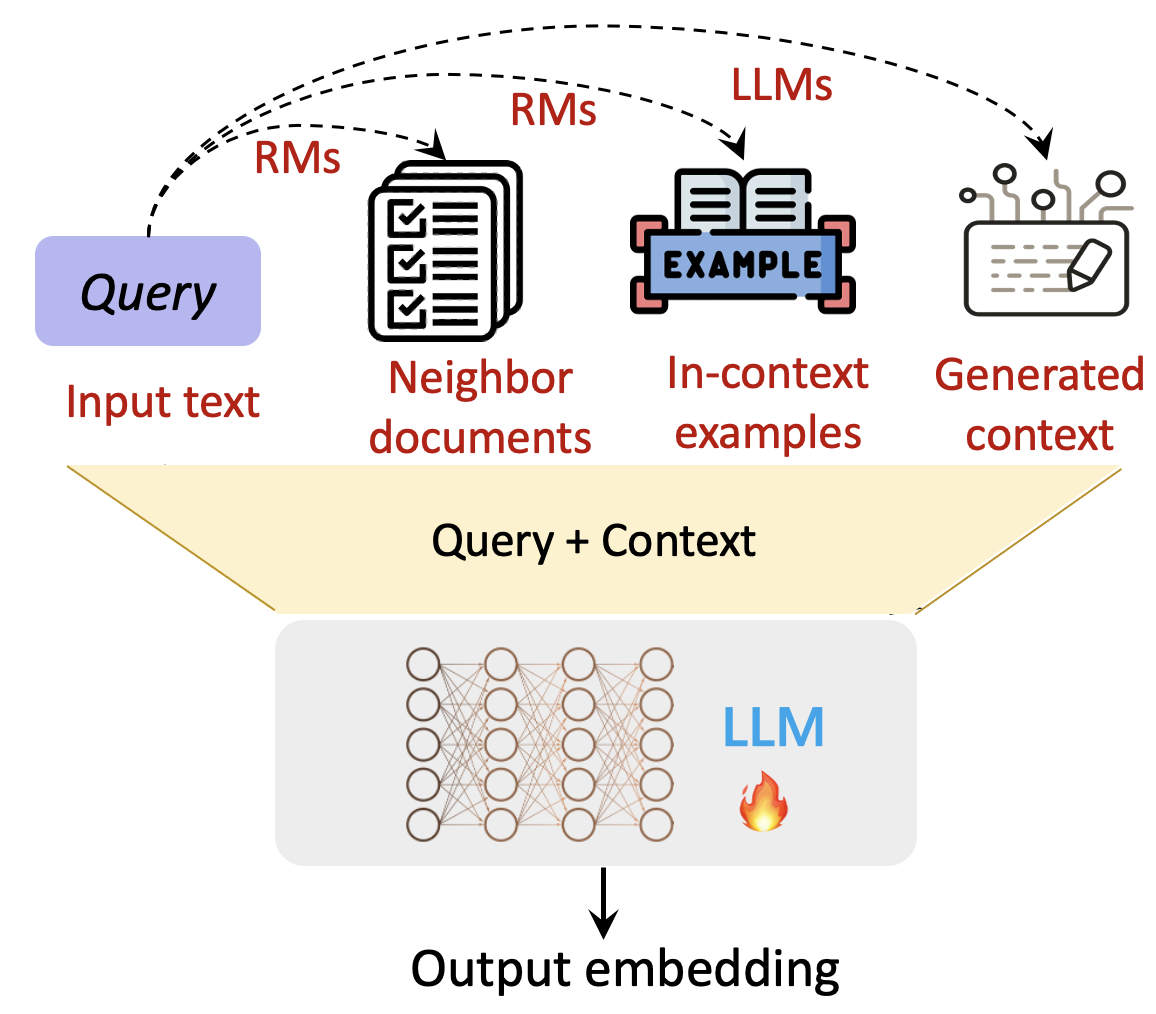}
\caption{An overview of contextual expansion for embedding models. RMs means retrieval models.}
\label{fig:Cont_emb}
\end{figure*}

\vspace{1mm}
\textbf{Contextual Expansion.} Compared with previous text embedding models which often only consider the text itself, some recent efforts expand the context of query or document to obtain enhanced text semantic information embedding.
\cite{li2024making} incorporated few-shot examples on the query side to enhance query embeddings during supervised contrastive training. 
\cite{morris2024contextuala} designed the alternative contrastive learning objective and the new contextual architecture to incorporate neighbor document information into document embeddings.
\cite{yan2025o1} generated thoughts towards the input query, then obtained the embedding of the query and generated thought separately, and finally incorporated both embeddings to produce the embedding that better reflects the query semantics.
\cite{ji2025learning} introduced the Chain-of-Deliberation mechanism, which encouraged LLM to encode documents with conducting reasoning process, enabling to generate more fine-grained document embeddings.
\cite{neeser2025quote} generated potential user queries based on each document, and through incorporating the embeddings of generated queries with the document embeddings, it's more effective to capture diverse user queries that reference the same document in different ways.
\cite{tejaswi2024rarea} proposed a simple yet effective method that enhances retriever models by training them with semantically similar in-context examples, enabling better utilization of contextual signals for improved retrieval performance.
\cite{li2025reinforced} proposed a Reinforced Information Retrieval framework in which the query-expansion model and embedding model mutually enhance each other. Expanded queries generate more informative embeddings, while refined embeddings guide more effective query expansions, creating a synergistic loop that improves retrieval accuracy. An overview of these contextual expansion techniques and their integration strategies within embedding-based retrieval models is illustrated in Figure~\ref{fig:Cont_emb}.


\begin{figure*}[ht!]
\centering
\includegraphics[width=0.7\linewidth]{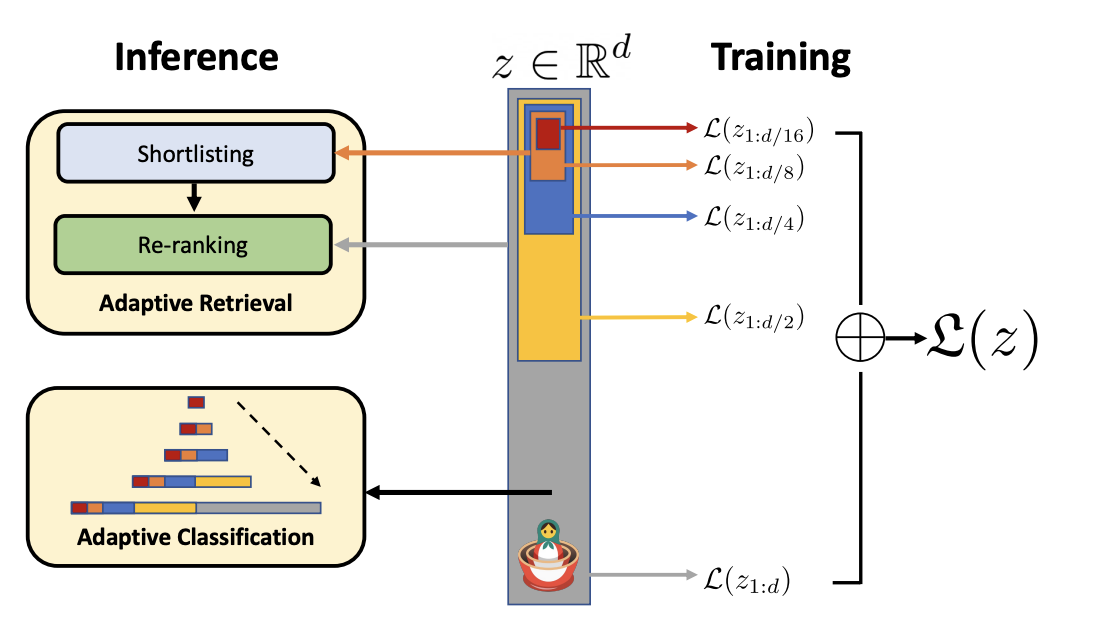}
\caption{An illustration of Matryoshka Embedding, from \cite{kusupati2022matryoshka}.}
\label{fig:mrl}
\end{figure*}

\vspace{1mm}
\textbf{Matryoshka Embedding (ME).}
Inspired by Russian nesting dolls, where smaller parts nest within larger ones. Matryoshka Embeddings \cite{kusupati2022matryoshka} are trained to store the most important information in the early dimensions of an embedding. This allows the embeddings to be truncated to smaller sizes (e.g., 64, 128, 256 dimensions) while still retaining useful information. As shown in Figure~\ref{fig:mrl}, by applying loss functions to multiple truncated versions during training, ME enables flexible trade-offs among performance, speed, and memory usage.
It’s especially useful for tasks like retrieval or classification, where quick filtering can be done with smaller embeddings, followed by re-ranking with full-size ones. More recently, The Qwen3 Embedding model series~\cite{zhang2025qwen3} employs the Matryoshka Embedding training approach and offers flexible, user-defined output dimensions ranging from 32 to 4096. This enables the model to better balance performance and computational efficiency across different applications.

\subsection{Training Objective}

Most existing methods for fine-tuning LLMs as embedding models adopt contrastive learning~\cite{chen2020simple}. The core idea behind contrastive learning is to bring representations of similar sequences closer in the embedding space while pushing dissimilar sequences apart, thereby learning robust and distinctive sequence embeddings. 
Formally, given the query $q_i$ and each of the candidate passage $p_j$, we can utilize a LLM to produces contextualized token embeddings for a sequence of tokens, then we can obtain the vector representations of \( q_i \) and \( p_i \) by taking the embedding of [EOS] token or averaging the token embeddings. To incorporate task-specific context, a task instruct \( t \) is prepended to each query.
The contrastive training process for the query \(q_i\) is formulated as the minimization of the following InfoNCE loss:
\begin{equation*}\label{eq:de-loss-sup} \small
    \mathcal{L} = - \log \frac{\exp^{\mathcal{S}(q_i, p^{+}_{i}) / \tau}}
    {\exp^{\mathcal{S}(q_i, p^{+}_{i}) / \tau} + 
    \sum\limits_{p^{-}_{i,j} \in \mathbb{P}^{-}_{i}} \exp^{\mathcal{S}(q_i, p^{-}_{i,j}) / \tau}}.
\end{equation*}
where $p_{i}^{+}$ is the labeled positive document paired with $q_i$ and $\mathbb{P}^{-}_{i}$ denotes the set of candidate documents for $q_i$ which is typically constructed during training by random negative sampling or hard negative mining methods. \(\mathcal{S}(q_i, p_i)\) denotes the similarity measure (e.g., cosine similarity) between the query and the positive. \(\tau\) is a temperature parameter that influences the sharpness of the similarity distribution.
By minimizing this contrastive loss, the model learns to bring embeddings of positive pairs closer together while pushing embeddings of negative pairs further apart. This process aids in creating more effective and discriminative text embeddings, enhancing the model's performance on a range of downstream tasks.

\vspace{1mm}
\textbf{Multi-task/stage Training.} 
Besides the standard contrastive objective, much effort has been made to improve the training procedures to enhance LLMs as versatile embedding models. 
For example, GTE~\cite{li2023towards} and Conan-embedding~\cite{li2024conanembedding} introduce a multi-stage training approach in which the model is first pretrained with an InfoNCE loss using in-batch negatives on weakly supervised text relevance data, followed by fine-tuning with a CoSENT loss on the STS task.
LLM2Vec~\cite{behnamghader2024llm2vec} transforms a pre-trained decoder-only LLM into a universal text encoder through \emph{masked next token prediction} and \emph{unsupervised contrastive learning}. This method uses only publicly available data and applies unsupervised contrastive training similarly to SimCSE~\cite{gao2021simcse}.
NV-Embed~\cite{lee2024nv} also focuses on training procedures to enhance LLMs. Specifically, the model first performs contrastive training with instructions on retrieval datasets and then integrates carefully curated non-retrieval datasets into the stage-one training data.
GritLM~\cite{muennighoff2024generative} proposes a unified model for both embedding tasks and generative tasks, which is jointly optimized with both NLL objective and contrastive loss.
\cite{zhang2025qwen3} also employs a multi-stage training pipeline, which includes weakly supervised training using large-scale synthetic paired data (over 150 million pairs), followed by supervised fine-tuning with high-quality synthetic and labeled data.

\vspace{1mm}
\textbf{Knowledge Distillation.} 
Additionally, several studies explore using knowledge distillation to enhance embedding performance by leveraging larger embedding models or cross-attention ranker models. Gecko~\cite{lee2024gecko}, for instance, builds a smaller bidirectional embedding model (with 1.2B parameters) by distilling knowledge from a decoder-only LLM by generating synthetic paired data. 
BGE-ICL~\cite{li2024making} introduce few-shot examples into the query side to enhance the query embeddings and also consider distilling the relevance score from the reranker for retrieval task during training.
\citet{zhang2023retrieve} introduce a unified embedding model designed to support the diverse retrieval augmentation needs of LLMs, including document knowledge, tools, in-context learning examples, and memory knowledge, through multi-task learning. The model incorporates a reward formulation based on LLM feedback and stabilizes knowledge distillation by integrating both soft reward-based labels and hard ranking-based labels during contrastive training.
\citet{lupart2024disco} propose distilling the scores of rewritten queries paired with documents from a teacher model and utilizing knowledge distillation from multiple teachers to enhance conversational representations in conversational search tasks.

\begin{figure*}[ht!]
\centering
\includegraphics[width=0.7\linewidth]{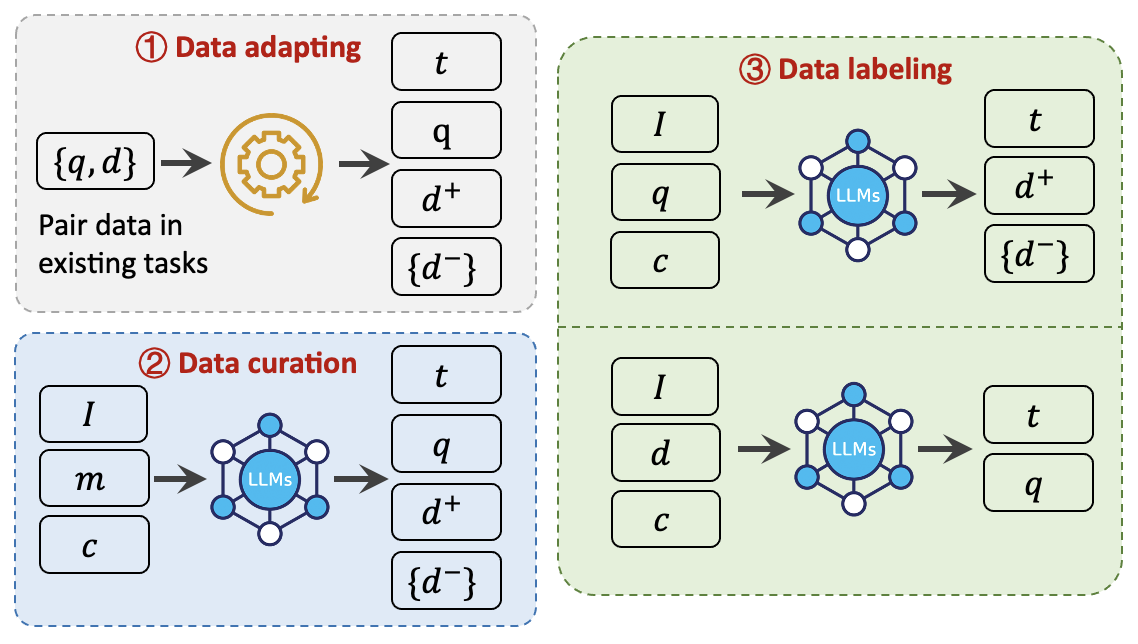}
\caption{An illustration of various data construction methods for training LLM-based embedding models. Here, $d^+$ denotes the positive document, and $\{d^-\}$ represents the set of hard negative documents. The variable $m$ refers to diverse meta-information used to guide synthetic data generation. $I$ is the instruction provided to the LLM for generating either a query $q$ or a document $d$, while $c$ optionally includes few-shot demonstrations. Finally, $t$ denotes the task-specific instruction.
}
\label{fig:data_aug}
\end{figure*}


\begin{table*}[ht!]
\small
\caption{Comparison of datasets used for supervised fine-tuning of several representative embedding models. QuestionRR means StackOverFlowDupQuestionRR. ``$-$" indicates that the model uses this type of data, but specific datasets are not specified.}

\setlength{\tabcolsep}{1.8pt}
\resizebox{1.0\linewidth}{!}{%
\begin{tabular}{cccccccccccc}
\toprule
TASK                                                  & Dataset                    & E5-Mistral                           & GritLM                       & Linq                         & SFR          & Gecko                & LLM2Vec                      & BGE-en-ICL                                       & NV-Embed                 &NV-Retriever & MTEB                                         \\ 
\midrule
  & DuReader                   & $\checkmark$                         & $\checkmark$                         & $\checkmark$                         & $\times$     & $\times$             & $\checkmark$                     &             $\times$                                 & $\times$                         & $\times$                         & $\times$                                     \\
  & ELI5                       & $\checkmark$                         & $\checkmark$                         & $\checkmark$                         & $\times$     & $\times$             & $\checkmark$                     &            $\checkmark$                                  & $\times$                         & $\times$                         & $\times$                                     \\
  & FEVER                      & $\checkmark$                         & $\checkmark$                         & $\checkmark$                         & $\checkmark$ & $\checkmark$         & $\checkmark$                     &           $\checkmark$                                   & $\checkmark$                     & $\checkmark$                         & $\checkmark$                                 \\
  & HotpotQA                   & $\checkmark$                         & $\checkmark$                         & $\checkmark$                         & $\checkmark$ & $\checkmark$         & $\checkmark$                     &        $\checkmark$                                      & $\checkmark$                     & $\checkmark$                         & $\checkmark$                                 \\
  & NLI                        & $\checkmark$                         & $\checkmark$                         & $\checkmark$                         & $\checkmark$ & $\checkmark$         & $\checkmark$                     &                                     $\checkmark$         & $\checkmark$                     & $\checkmark$                         & $\times$                                     \\
  & MIRACL                     & $\checkmark$                         & $\checkmark$                         & $\checkmark$                         & $\times$     & $\checkmark$         & $\checkmark$                     &      $\times$                                        & $\times$                         & $\times$                         & $\times$                                     \\
  & MSMARCO                    & $\checkmark$                         & $\checkmark$                         & $\checkmark$                         & $\checkmark$ & $\times$             & $\checkmark$                     &                                       $\checkmark$       & $\checkmark$                     & $\checkmark$                         & $\checkmark$                                 \\
  & Mr.TyDi                    & $\checkmark$                         & $\checkmark$                         & $\checkmark$                         & $\times$     & $\times$             & $\checkmark$                     &                 $\times$                              & $\times$                         & $\times$                         & $\times$                                     \\
  & NQ                         & $\checkmark$                         & $\checkmark$                         & $\checkmark$                         & $\checkmark$ & $\checkmark$         & $\checkmark$                     &                                        $\checkmark$      & $\checkmark$                     & $\checkmark$                     & $\checkmark$                                 \\
  & S2ORC                      & $\times$                             & $\checkmark$                         & $\checkmark$                         & $\times$     & $\times$             & $\times$                         &   $\times$                                           & $\times$                         & $\times$                         & $\times$                                     \\
  & SQuAD                      & $\checkmark$                         & $\checkmark$                         & $\checkmark$                         & $\times$     & $\times$             & $\checkmark$                     &          $\checkmark$                                    & $\checkmark$                     & $\checkmark$                     & $\times$                                     \\
  & T2Ranking                  & $\checkmark$                         & $\checkmark$                         & $\checkmark$                         & $\times$     & $\times$             & $\checkmark$                     &  $\times$                                            & $\times$                         & $\times$                         & $\times$                                     \\
  & TriviaQA                   & $\checkmark$                         & $\checkmark$                         & $\checkmark$                         & $\times$     & $\times$             & $\checkmark$                     &         $\checkmark$                                     & $\times$                         & $\checkmark$                         & $\times$                                     \\
  & Quora                      & $\checkmark$                         & $\checkmark$                         & $\checkmark$                         & $\checkmark$ & $\times$             & $\checkmark$                     &                  $\checkmark$                            & $\times$                         & $\times$                         & $\checkmark$                                 \\
  & FiQA                       & $\times$                             & $\times$                             & $\times$                             & $\checkmark$ & $\times$             & $\times$                         &          $\checkmark$                        & $\checkmark$                     & $\checkmark$                         & $\checkmark$                                 \\
  & SciFact                    & $\times$                             & $\times$                             & $\times$                             & $\checkmark$ & $\times$             & $\times$                         &              $\times$                                & $\times$                         & $\checkmark$                         & $\checkmark$                                 \\
  & NFCorpus                   & $\times$                             & $\times$                             & $\times$                             & $\checkmark$ & $\times$             & $\times$                         &         $\times$                                     & $\times$                         & $\checkmark$                         & $\checkmark$                                 \\
  & DBPedia                    & $\times$                             & $\times$                             & $\times$                             & $\checkmark$ & $\times$             & $\times$                         & $\times$                         & $\times$                         & $\times$                         & $\checkmark$                                 \\
  & MedMCQA                    & $\times$                             & $\times$                             & $\times$                             & $\times$     & $\checkmark$         & $\times$                         &        $\times$                              & $\times$                         & $\times$                         & $\times$                                     \\
  & PAQ                        & $\times$                             & $\times$                             & $\times$                             & $\times$     & $\times$             & $\times$                         & $\times$                        & $\checkmark$                     & $\checkmark$                         & $\times$                          \\
  & Stackexchange              & $\times$                             & $\times$                             & $\times$                             & $\times$     & $\times$             & $\times$                         & $\times$                         & $\checkmark$                     & $\checkmark$                     & $\checkmark$                         \\
  & ArguAna                    & $\times$                             & $\times$                             & $\times$                             & $\times$     & $\times$             & $\times$                         & $\checkmark$                         & $\checkmark$                     & $\checkmark$                         & $\checkmark$                            \\
  & GOOAQ & $\times$                             & $\times$                             & $\times$                             & $\times$     & $\times$             & $\times$                         & $\times$                         & $\times$                     & $\checkmark$                         & $\times$                            \\
\multirow{-24}{*}{Retrieval}                          & BioASQ                     & $\times$                             & $\times$                             & $\times$                             & $\times$     & $\times$             & $\times$                         & $\times$                         & $\checkmark$                     & $\checkmark$                         & $\times$                         \\
        \midrule
\rowcolor[HTML]{CBCEFB} 
\cellcolor[HTML]{CBCEFB}                              & arXiv                      & $\times$                             & $\times$                             & $\times$                             & $\checkmark$ & $\times$             & \cellcolor[HTML]{CBCEFB}$\times$ &                 $\checkmark$                             & $\times$                         & \cellcolor[HTML]{CBCEFB}$\checkmark$ & $\checkmark$                                 \\
\rowcolor[HTML]{CBCEFB} 
\cellcolor[HTML]{CBCEFB}                              & Reddit                   & $\times$                             & $\times$                             & $\times$                             & $\times$ & $\times$             & \cellcolor[HTML]{CBCEFB}$\times$ &                  $\checkmark$                            & $\times$                         & \cellcolor[HTML]{CBCEFB}$\times$ & $\checkmark$                                 \\

\rowcolor[HTML]{CBCEFB} 
\cellcolor[HTML]{CBCEFB}                              & StackExchange                   & $\times$                             & $\times$                             & $\times$                             & $\times$ & $\times$             & \cellcolor[HTML]{CBCEFB}$\times$ &                  $\checkmark$                            & $\times$                         & \cellcolor[HTML]{CBCEFB}$\times$ & $\checkmark$                                 \\

\rowcolor[HTML]{CBCEFB} 
\cellcolor[HTML]{CBCEFB}                              & bioRxiv                    & $\times$                             & $\times$                             & $\times$                             & $\checkmark$ & $\times$             & \cellcolor[HTML]{CBCEFB}$\times$ &                  $\checkmark$                            & $\times$                         & \cellcolor[HTML]{CBCEFB}$\checkmark$ & $\checkmark$                                 \\
\rowcolor[HTML]{CBCEFB} 

\multirow{-3}{*}{\cellcolor[HTML]{CBCEFB}Clustering}  & medRxiv                    & $\times$                             & $\times$                             & $\times$                             & $\checkmark$ & $\times$             & \cellcolor[HTML]{CBCEFB}$\times$ &                  $\checkmark$                            & $\times$                         & \cellcolor[HTML]{CBCEFB}$\checkmark$ & $\checkmark$                                 \\
\midrule
  & AmazonReview               & $\times$                             & $\times$                             & $\times$                             & $\checkmark$ &  -                    & $\times$                         &     $\checkmark$                                         & $\checkmark$                     & $\checkmark$                         & $\checkmark$                                 \\
  & Emotion                    & $\times$                             & $\times$                             & $\times$                             & $\checkmark$ & -                     & $\times$                         &      $\checkmark$                                        & $\checkmark$                     & $\checkmark$                         & $\checkmark$                                 \\
  & MTOPIntent                 & $\times$                             & $\times$                             & $\times$                             & $\checkmark$ &     -                 & $\times$                         &                  $\checkmark$                            & $\checkmark$                     & $\checkmark$                         & $\checkmark$                                 \\
  & ToxicConversation          & $\times$                             & $\times$                             & $\times$                             & $\checkmark$ &     -                 & $\times$                         &      $\checkmark$                                        & $\checkmark$                     & $\checkmark$                         & $\checkmark$                                 \\
  & TweetSentiment             & $\times$                             & $\times$                             & $\times$                             & $\checkmark$ &  -                    & $\times$                         &         $\checkmark$                                     & $\checkmark$                     & $\checkmark$                         & $\checkmark$                                 \\
  & AmazonCounterfactual       & $\times$                             & $\times$                             & $\times$                             & $\times$     & - & $\times$                         & $\checkmark$                         & $\checkmark$                     & $\checkmark$                         & $\checkmark$                          \\
  & Banking77                  & $\times$                             & $\times$                             & $\times$                             & $\times$     & - & $\times$                         & $\checkmark$                         & $\checkmark$                     & $\checkmark$                         & $\checkmark$                          \\
\multirow{-8}{*}{Classification}                      & IMDB                       & $\times$                             & $\times$                             & $\times$                             & $\times$     &            -          & $\times$                         &                 $\checkmark$                             & $\checkmark$                     & $\checkmark$                         &           $\checkmark$                                    \\
        \midrule
\rowcolor[HTML]{CBCEFB} 
\cellcolor[HTML]{CBCEFB}                              & STS12                      & $\times$                             & $\times$                             & $\times$                             & $\checkmark$ & $\times$             & \cellcolor[HTML]{CBCEFB}$\times$ &                                   $\checkmark$           & $\times$                         & \cellcolor[HTML]{CBCEFB}$\checkmark$ & $\checkmark$                                 \\
\rowcolor[HTML]{CBCEFB} 
\cellcolor[HTML]{CBCEFB}                              & STS22                      & $\times$                             & $\times$                             & $\times$                             & $\checkmark$ & $\times$             & \cellcolor[HTML]{CBCEFB}$\times$ &                           $\checkmark$                   & $\times$                         & \cellcolor[HTML]{CBCEFB}$\checkmark$ & $\checkmark$                                 \\
\rowcolor[HTML]{CBCEFB} 
\multirow{-3}{*}{\cellcolor[HTML]{CBCEFB}STS}         & STSBenchmark               & $\times$                             & $\times$                             & $\times$                             & $\checkmark$ & $\times$             & \cellcolor[HTML]{CBCEFB}$\times$ &                   $\checkmark$                           & $\times$                         & \cellcolor[HTML]{CBCEFB}$\checkmark$ & $\checkmark$                                 \\

\midrule

& SciDocs                    & $\times$                             & $\times$                             & $\times$                             & $\checkmark$ & $\times$             & $\times$                         &          $\checkmark$                                    & $\times$                         & $\times$                         & $\checkmark$                                 \\
\multirow{-2}{*}{Reranking}                           & QuestionRR & $\times$                             & $\times$                             & $\times$                             & $\checkmark$ & $\times$             & $\times$                         &                         $\checkmark$                     & $\times$                         & $\times$                         & $\checkmark$                                 \\

\midrule
\midrule

Synthesic & -      &  $\checkmark$ &  $\checkmark$ &  $\checkmark$ & $\times$     & $\checkmark$         & $\times$ &  $\times$ &  $\times$ &  $\times$ &  $\times$\\

\bottomrule
\end{tabular}
}
\label{tab:data}
\end{table*}

\subsection{Data Construction}

Training data plays a crucial role in the success of fine-tuning LLMs as embedding models.The methods typically employ two types of data: (1) adapting existing datasets, and (2) synthesizing datasets via LLMs. The latter often involves both data curation and automated labeling to construct training pairs. Figure~\ref{fig:data_aug} illustrates representative methods for constructing contrastive training datasets under these paradigms.

\vspace{1mm}
\textbf{Adapting Existing Datasets.}
Table~\ref{tab:data} presents the most commonly used public datasets, which include retrieval, clustering, classification, STS, and reranking tasks that align with the evaluation criteria in the Massive Text Embedding Benchmark (MTEB)\footnote{\url{https://huggingface.co/spaces/mteb/leaderboard}}~\cite{muennighoff2022mteb}. These datasets provide diverse labeled examples that can be used to fine-tune models in a supervised manner, enhancing their ability to capture specific semantic relationships and improve performance on downstream embedding tasks.
Specifically, these tasks are typically divided into symmetric and asymmetric tasks~\cite{wang2023improving}. Symmetric tasks consist of queries and documents that convey similar meanings but differ in their wording, while asymmetric tasks include tasks where the query and document share a semantic relationship without being direct paraphrases.
Besides paired data, some works leverage self-supervised learning methods like SimCSE, using natural language text from sources such as Wikipedia~\cite{behnamghader2024llm2vec}. These tasks often involve generating different views of the same input sequence through augmentations, such as dropout or token masking, to create positive pairs for contrastive learning. 
Moreover, SFR-Embedding-Mistral~\cite{meng2024sfrembedding} fine-tunes the E5 model across a diverse array of tasks, and demonstrates that the effectiveness of embedding models can be enhanced through knowledge transfer from multiple tasks.

\vspace{1mm}
\textbf{Synthesizing Datasets via LLMs}
Considering the limited quantity of supervised data and the uneven distribution of corresponding pair relevance, recent research has addressed this challenge by leveraging LLMs to automatically synthesize large-scale, diverse, and contextually rich datasets~\cite{wang2023improving}. This approach draws inspiration from LLM-based data augmentation in the supervised fine-tuning process~\cite{xu2024survey}.
Synthetic datasets for training embedding models typically involve two primary methods: data labeling and data curation.  In the labeling approach, a query is provided for the LLM to generate a relevant document as a positive example\cite{wang2023improving}, optionally producing hard negative documents\cite{wang2023improving}, or to generate a query\cite{zhang2025qwen3,lee2024gecko} for a sampled document from the corpus.  In the curation method, the LLM is prompted to generate both the query and the relevant document, utilizing few-shot query-document examples as optional information. By using model-generated data, researchers can produce large-scale, high-quality training datasets tailored to specific embedding tasks, improving the robustness and generalizability of the fine-tuned models.

In the early stage of LLM area, \cite{jeronymo2023inpars,jeronymo2023inparsv2} utilize GPT-J to generate one synthetic query per given document, prompted with three examples from MS MARCO, and then leverage a powerful reranker to select the synthetic query-document pairs for training. E5-Mistral \cite{wang2023improving} introduce a two-step prompting method in which proprietary LLMs (GPT-4) are initially prompted to create a pool of candidate tasks. From this pool, a specific task is chosen, and the LLMs are prompted again to generate triples consisting of a query, a positive document, and a hard-negative document. This method ultimately supports a broad range of text embedding tasks across 93 languages, covering hundreds of thousands of embedding tasks.
Gecko~\cite{lee2024gecko} prompted LLMs to read a sampled passage from a web corpus to generate both a task description and a relevant query, after which the top-K relevant passages were retrieved for these synthetic queries, and LLMs were employed to relabel more relevant positive passages along with a better hard negative.
Linq-Embed-Mistral~\cite{kim2024linq} utilized GPT-4-turbo to generate the query-positive-negative triplet as training samples across six tasks, where the synthetic strategy was closely adhering to the \cite{wang2023improving}. Additionally, \cite{kim2024linq} conducted a data refinement on synthetic data to improve overall data quality through employing various types of few-shot prompt engineering, filtering, and negative mining.
\cite{khramtsova2024leveraging} explored the zero-shot capabilities of LLMs for generating queries, relevance judgments, and reference lists. 
\cite{lee2025gemini} leveraged multi-stage prompting strategies to generate training samples across retrieval and classification tasks. For retrieval, Gemini was prompted to generate queries for web passages followed by a Gemini auto-rater to filter lower-quality examples and for classification, counterfactual, sentiment, and review classification training samples were generated. 
\cite{zhang2025qwen3} prompted Qwen3-32B to generate queries for the documents by employing diverse prompting strategies, where specific roles were assigned to each document with incorporating various generative dimensions including query type, length, difficulty, and language.

\vspace{1mm}
\textbf{Negative Mining.}
Negative mining is crucial for constructing high-quality training data for embedding models, as appropriate negative samples enhance the representation capacity of embedding models trained with contrastive loss. Consistent with previous representation learning methods~\cite{karpukhin2020dense,xiong2020approximate}, current LLM-based embedding models are exploring various methods to mine negative to improve the training process. A naive negative mining technique is using positive document examples from other queries within the same training batch as negatives~\cite{chen2020simple}. While efficient, this method often yields overly simple and uninformative negatives, which may hinder effective contrastive learning.
To address this limitation, recent works focus on constructing hard negatives using two primary methods: selection and generation. 

The selection method involves identifying hard negatives from the candidate corpus. \cite{su2023one} employed Sentence-T5 embeddings~\cite{ni-etal-2022-sentence} to encode pairs of texts, calculating pairwise cosine similarity scores to determine suitable hard negatives. 
\cite{lee2024gecko} retrieved the top-K relevant passages for synthetic queries and utilized LLMs to relabel more relevant positive passages and identify better hard negatives.
\cite{li2024conanembedding} introduce dynamic hard negative mining, which selects more challenging negative examples during contrastive training.
During the selection process, some studies have noted the issue of false negatives, where certain hard negatives should actually be classified as positive examples~\cite{qu2020rocketqa}. To address this, \cite{moreira2024nv} examined various filtering methods for hard negatives, proposing the use of similarity scores from positive examples as thresholds with specific margins or percentages.

In contrast, the generation method leverages LLMs to directly create hard negatives for queries. 
\cite{wang2023improving} utilized LLMs to generate query, positive, hard-negative triples based on a selected task from pre-generted candidate tasks pool.
This synthetic strategy has been applied to various embedding models~\cite{muennighoff2024generative,springer2024repetition,kim2024linq}, demonstrating its versatility.
Building on the approach of \cite{wang2023improving}, \cite{kim2024linq} introduced a data refinement process, such as such as few-shot prompt engineering, filtering, and improved negative mining, to enhance the quality of synthetic data. 
This integration of hard negative mining techniques directly exploits the powerful generative capabilities of LLMs, ultimately facilitates more robust representational learning, enhancing the capabilities of embedding models.

\section{Specialized Embeddings}
Beyond regular text embeddings, advanced techniques have been developed to handle longer texts, multiple languages, cross-modal data, and programming code. This section delves into these specialized embedding techniques, highlighting their methodologies and applications.

\subsection{Multi-lingual Embedding}

Multi-lingual embeddings are designed to represent text from different languages in a shared vector space. These embeddings are useful for various tasks such as cross-lingual information retrieval, machine translation, and multi-lingual sentiment analysis. The development of monolingual language models has substantially progressed in learning embeddings enriched with contextual information across a range of specific languages.

Early pretrained LMs like BERT ~\cite{devlin2018bert} and GPT ~\cite{radford2018improving} learn fine-grained contextual monolingual representations through masked language modeling (MLM) and next-token prediction (NTP). Built on the foundation of monolingual LMs, a series of advanced works have been proposed to learn universal multi-lingual embeddings. 
We categorize these approaches based on their alignment techniques into two types: \emph{implicit alignment} and \emph{explicit alignment}.

\subsubsection{Implicit alignment.}
Instead of direct translations or aligned sentence pairs, implicit alignment leverages shared representations and latent relationships within the training data. By utilizing techniques such as shared vocabularies, cross-lingual embeddings, and self-supervised learning, multilingual LLMs can well capture semantic similarities and syntactic structures across diverse languages and implicitly align different languages into a unified semantic embedding space. This approach enhances the model's ability to perform tasks in low-resource languages, where paired data may be scarce, ultimately leading to more robust and versatile multilingual capabilities.

Multi-lingual Language Models such as mBERT ~\cite{pires2019multilingual}, XLM-R ~\cite{conneau2019unsupervised}, mT5 ~\cite{xue2020mt5} have explored the capabilities of language models on multilingual NLP tasks. Recently, with model parameters and training data scaled up, numerous multilingual large language models (MLLMs), such as GPT-3 ~\cite{brown2020language}, Gopher ~\cite{rae2021scaling}, LaMDA ~\cite{thoppilan2022lamda}, InstructGPT ~\cite{ouyang2022training}, PaLM ~\cite{chowdhery2023palm}, BLOOM ~\cite{workshop2022bloom}, LLaMA ~\cite{touvron2023llama}, PaLM 2 ~\cite{anil2023palm}, LLaMA 2 ~\cite{touvron2023llama2}, GLM-130B ~\cite{zeng2022glm}, have achieved impressive multilingual performance. 

Multilingual capabilities learned through implicit alignment in MLMs can be evidenced by some works that use extracted multilingual embeddings to complete downstream multilingual tasks.
For instance, M3-Embedding~\cite{chen2024bge}, based on XLM-RoBERTa~\cite{conneau2019unsupervised} and trained on a large-scale, diverse multilingual dataset, employs self-distillation to address multilingual hybrid retrieval tasks, including dense retrieval, lexical (sparse) retrieval, and multi-vector retrieval. 
Additionally, to overcome the limitations of existing embedding models in Malay retrieval tasks,
the Multi-Lingual Malaysian Embedding~\cite{zolkepli2024multi} pre-trained and fine-tuned Malaysian Llama2 on collected synthetic Malaysian RAG and hard mining datasets. Features from different hidden layers of the Malaysian Llama2 were extracted to cater to the needs of various application scenarios.

\cite{sturua2024jinaembeddingsv3} introduced jina-embeddings-v3, a multilingual embedding model employs task-specific Low-Rank Adaptation (LoRA) adapters to enhance performance across various downstream tasks. The model supports long-context retrieval with sequences up to 8192 tokens by integrating RoPE~\cite{su2024roformer}. Additionally, jina-embeddings-v3 incorporates Matryoshka Representation Learning~\cite{mrl} to allow flexible reduction of embedding dimensions without compromising performance, enabling embeddings to be truncated to as low as 32 dimensions. By addressing common retrieval failures using synthetic data and leveraging instruction tuning~\cite{wei2021finetuned,1emb2anytask}, the model achieves state-of-the-art results on the MTEB benchmark~\cite{muennighoff2022mteb}, outperforming proprietary embeddings from OpenAI and Cohere on English tasks and surpassing multilingual-e5-large-instruct~\cite{wang2022text} across all multilingual tasks.

\subsubsection{Explicit Alignment.}
Unfriendly to low-resource languages, implicit alignment often requires training a multilingual model from scratch, which typically demands extensive computational costs and multilingual data. Specifically, both the large corpus used for pretraining and the fine-grained data for supervised fine-tuning (SFT) or reinforcement Learning with Human Feedback (RLHF) are hard to obtain. MLLMs tend to perform poorly on low-resource languages due to the imbalance in the corpus. Therefore, recent work has focused on bootstrapping multilingual embedding learning with enrich multilingual representations from pre-trained models, such as monolingual expert language models or machine translation models.

MT-LLM ~\cite{schmidt2024self} aligns the multilingual representations, which are extracted by machine translation encoder, into the semantic embedding space of LLMs via self-distillation. This integration enables the LLMs to perform zero-shot ability to any language supported by the machine translation encoder.  
\cite{vasilyev2024preserving} employs a methodology that exclusively fine-tunes the query encoder while keeping the text encoder frozen on an english-only dataset, finding that this approach not only preserves but significantly enhances the multilingual embedding capabilities of the model.
Based on the assumption that monolingual embeddings are well-structured, \cite{Park2024ImprovingMA} proposes to distill alignment information from the monolingual similarity matrix into cross-modal embeddings to guide the cross-lingual alignment process. 
This method resolves issues in prior contrastive learning approaches that treated non-exact translation pairs as negative samples, which disrupted the monolingual embedding space.
In addition, to further assess the capabilities of multilingual language models, ~\cite{yoon2024langbridge} propose MINERS, a multilingual, multi-task, tuning-free benchmark specifically designed to evaluate semantic retrieval tasks including bitext mining and classification via retrieval-augmented contexts.


\begin{table}[h!]
\centering
\caption{Recent Benchmarks for Evaluating Multilingual and Language-Specific.}
\resizebox{0.8\linewidth}{!}{
\begin{tabular}{llc}
\toprule
\textbf{Benchmark} & \textbf{Language Coverage} & \textbf{Year} \\
\midrule
MMTEB~\cite{enevoldsen2025mmteb} & 250+ languages & 2025 \\
mFollowIR~\cite{weller2025mfollowir} & Persian, Chinese, Russian & 2025 \\
FaMTEB~\cite{zinvandi2025famteb}  & Persian & 2025 \\
C-MTEB~\cite{xiao2024c}  & Chinese & 2024 \\
PL-MTEB~\cite{poswiata2024pl}   & Polish & 2024 \\
Scandinavian Embedding Benchmark (SEB)~\cite{enevoldsen2024scandinavian}  & Nordic languages & 2024 \\
ArabicMTEB~\cite{bhatia2024swan}  & Arabic dialects, cross-lingual & 2024 \\
RusBEIR~\cite{kovalev2025building} & Russian & 2024 \\
JMTEB~\cite{jmteb}& Japanese & 2024 \\
Mteb-french~\cite{ciancone2024mteb} & French & 2024 \\
Amharic Passage Retrieval Benchmark~\cite{amde2025optimized} & Amharic & 2025 \\
\bottomrule
\end{tabular}
}
\label{tab:multilingual_benchmarks}
\end{table}

\subsubsection{Multilingual Text Embedding Benchmark}
Recent efforts have also focused on benchmarking multilingual embeddings to evaluate and advance their effectiveness. Among these, the Massive Multilingual Text Embedding Benchmark (MMTEB)~\cite{enevoldsen2025mmteb}, an extension of the well-known MTEB~\cite{muennighoff2022mteb}, provides a comprehensive evaluation across over 250 languages and 500 tasks spanning 10 categories. In addition to MMTEB, specialized multilingual benchmarks have been proposed for specific languages or language groups, such as mFollowIR~\cite{weller2025mfollowir} for instruction-following retrieval, FaMTEB~\cite{zinvandi2025famteb} for Persian, C-MTEB~\cite{xiao2024c} for Chinese, PL-MTEB~\cite{poswiata2024pl} for Polish, the Scandinavian Embedding Benchmarks~\cite{enevoldsen2024scandinavian} covering Nordic languages, Swan and ArabicMTEB~\cite{bhatia2024swan} focusing on Arabic dialects and cross-cultural embeddings, as well as JMTEB~\cite{jmteb} for Japanese and various benchmarks targeting German~\cite{wehrli2024german}, targeting French~\cite{ciancone2024mteb}, Russian~\cite{kovalev2025building}, and Amharic~\cite{amde2025optimized}. These benchmarks collectively address the linguistic diversity, dialectal variations, and cultural contexts inherent in multilingual NLP, providing crucial standardized frameworks for evaluating embedding quality and facilitating progress in universal and language-specific embedding models.

\subsection{Code Embedding}
LLM-based code embeddings transform programming code into continuous vector representations that capture semantic, syntactic, functional properties and logic rules. These embeddings enable retrieval systems to encode code similarly to natural language, powering various types of code retrieval tasks.
Traditional text embedding approaches struggle with code retrieval tasks which is largely due to a mismatch between the semantic structures of code and natural language. Therefore, it's important to align features within and between code and natural languages.
Early code embedding works like code2vec~\cite{alon2019code2vec} and Codebert~\cite{feng2020codebert}, adapted NLP techniques to code by treating code as text. However, these methods struggled with structural nuances (e.g., control flow, data dependencies) and multi-language support.
GraphCodeBERT~\cite{guo2020graphcodebert} improved this by incorporating abstract syntax trees (ASTs). Further, UniXcoder~\cite{guo2022unixcoder} unifies code representations (tokens, ASTs, comments) into a single multimodal transformer, enabling diverse code understanding and generation tasks like code search and summarization.

With the development of LLMs, some LLM-based code embedding models were proposed to alleviate the limitation. 
~\cite{li2023starcoder} released StarCoderBase and StarCoder models, trained on large amounts of code data across more than 80 programming languages which were sourced from GitHub issues, Git commits, and Jupyter notebooks. They also used multi-query attention for efficient long-context embedding (8K tokens).
~\cite{liu2024codexembed} proposed a generalizable training framework which converted diverse code-related tasks into retrieval tasks, and constructed training samples from code-to-text, text-to-code, code-to-code, text-to-text and hybrid text and code tasks across multiple programming languages.
~\cite{li2025towards} introduced CodeR trained on a large-scale synthetic training dataset, which was created via a novel pipeline using LLMs for task design and sample generation. Meanwhile they employed a three-stage curriculum learning strategy to optimize training process.

\subsection{Long Context Embedding}
The ability to effectively capture and utilize long context information has become increasingly vital. Traditional embedding techniques, which typically focus on limited length of text, often struggle to maintain coherence and capture the full semantic meaning when applied to texts with extended contexts. Long context embeddings are designed to address these challenges, enabling models to handle extended sequences of text while preserving contextual integrity and meaning.
A popular method for extending the length of context is training the embedding models with the long-context backbone, which can be obtained either by using the existing model or by pre-training with long inputs from scratch~\cite{wang2023improving,nussbaum2024nomic,gunther2023jina,chen2024bge}.
Typically, \cite{zhang2024mgte} introduced a text encoder enhanced with RoPE~\cite{su2024roformer} and unpadding~\cite{portes2024mosaicbert}, which is pre-trained by masked language model based on much longer (e.g. 8K tokens) and multilingual context with a two-stage curriculum. Based on the text encoder, they construct the embedding model through contrastive pre-training and finetuning utilizing InfoNCE as the loss function.

However, fine-tuning LLMs may result in high computational costs, where there are more and more efforts are proposed to extend the contexts of LLM embedding models through the plug-and-play methods~\cite{ratner2022parallel,jin2024llm,wang2024resonance}.
\cite{luo2024bge} proposes a chunking-free architecture to process long context, which adds a special token <LMK> at the end of each sentence in the long context to capture the coherent semantic. During training, this work introduces the positional weight and utilizes multi-stage training to achieve a superior performance.
\cite{shao2024extensible} proposes a plug-in module extensible embedder to process the long context which is partitioned into multiple chunks. Each chunk is embedded and down-scaled by extensible embedder as the compact representation. This work train extensible embedder through two-stream auto-regression with fixed downstream LLM’s parameters, which does not affect the LLM’s original capabilities.
Furthermore, \cite{zhu2024longembed} explores the extensive plug-and-play strategies to extend existing embedding models to long context, which includes parallel context windows, grouped positions, recurrent positions, linear position interpolation and so on. 
This work also proposes a newly constructed LONGEMBED benchmark for long context retrieval evaluation, which includes two synthetic tasks with flexible document length and four real tasks with featuring dispersed target information.

\subsection{Cross-modal Embedding}
Cross-modal embeddings aim to create a unified representation for data from different modalities, such as text, images, and audio. These embeddings enable the integration of multimodal information, which is crucial for applications like image captioning, visual QA, and multimodal search.

The potential of Transformers for efficiently learning cross-modal embeddings has been demonstrated by works like VATT ~\cite{akbari2021vatt}, ViT ~\cite{dosovitskiy2020image}. Following the introduction of CLIP ~\cite{radford2021learning}, which uses language as supervision for visual models to bridge the gap between vision and language, a surge of vision-language cross-modal models such as ALBEF ~\cite{li2021align}, VLMO ~\cite{bao2021vlmo}, CoCa ~\cite{yu2022coca}, BLIP ~\cite{li2022blip}, and BEiT ~\cite{bao2021beit} emerged.
However, these models have significant limitations in representing purely textual or purely visual data. The VISTA ~\cite{zhou2024vista} proposes a new embedding approach for universal multimodal retrieval, allowing the pretrained LMs to recognize image tokens by utilizing ViT as an image tokenizer. It undergoes two-stage training on datasets with weak labels and high-quality synthetically composed image-text datasets, achieving superior performance across a variety of multimodal retrieval tasks.

With the development of LLMs, many efforts have extended LLMs to various modalities, such as image (BLIP-2 ~\cite{li2023blip}, MiniGPT4 ~\cite{zhu2023minigpt}, Llava ~\cite{liu2024visual}), audio (AudioPaLM ~\cite{rubenstein2023audiopalm}, Qwen-Audio ~\cite{chu2023qwen}), and video (Video-Llama ~\cite{zhang2023video}, Video-ChatGPT ~\cite{maaz2023video}, Video-Llava ~\cite{lin2023video}). However, these efforts primarily focus on generating text related to these modalities.
PaLM 2 DE ~\cite{gomez2024transforming} incorporates LLMs into dual-encoder architecture, utilizing PaLM to initialize and augment multilingual text comprehension for cross-modal retrieval tasks. This integration results in enhanced performance across an impressive range of 102 languages, even though it was trained on a limited dataset of 21 languages. Furthermore, by leveraging machine translation data, PaLM 2 DE significantly boosts its cross-lingual capabilities.

\begin{table}[ht!]
\centering
\caption{Summary of recent reasoning-based retrieval benchmarks.}

\renewcommand{\arraystretch}{1.2}
\resizebox{\linewidth}{!}{%
\begin{tabular}{p{3cm} p{1cm} p{3.2cm} p{5.5cm} p{2.4cm}}
\toprule
\textbf{Dataset} & \textbf{Year} & \textbf{Task Type} & \textbf{Reasoning Required} & \textbf{Domain} \\
\midrule
\textbf{RAR-b}~\cite{xiao2024rar} & 2024 & Reasoning-Intensive Retrieval  & temporal, numerical, spatial, symbolic & General \\
\textbf{ImpliRet}~\cite{taghavi2025impliret} & 2025 & Reasoning-Intensive Retrieval  & arithmetic, temporal, commonsense & General \\
\textbf{BRIGHT}~\cite{su2025bright} & 2025 & Reasoning-Intensive Retrieval  & Symbolic, math, code & General \\

\midrule
\textbf{InstructIR}~\cite{oh2024instructira} & 2024 & Instructional IR & Instruction following & General \\

\textbf{InfoSearch}~\cite{zhou2025contenta} & 2024 & Instructional IR &  Customized instruction following & General \\
\textbf{FollowIR}~\cite{weller2024followira} & 2024 & Instructional IR & Long-Form instruction comprehension & General \\
\textbf{MAIR}~\cite{sun2024maira} & 2024 & Instructional IR & Diverse instruction following & General \\

\textbf{BIRCO}~\cite{wang2024birco} & 2024 & Instructional IR & Compositional, Complex Objectives & General \\

\textbf{mFollowIR}~\cite{weller2025mfollowir} & 2025 & Instructional IR & Multi-lingual instruction following & General \\
\textbf{IFIR}~\cite{song2025ifir} & 2025 & Instructional IR & Domain-specific instruction following & Biomedical, Legal, Technical \\

\midrule
\textbf{CLERC}~\cite{hou2024clerc} & 2024 & Case/Legal Retrieval & Precedent Reasoning, Legal Logic & Legal \\
\textbf{Bar Exam/ Housing Statute}~\cite{zheng2025reasoningfocused} & 2025 & Case/Legal Retrieval & Symbolic, Legal Reasoning & Legal \\
\bottomrule
\end{tabular}%
}
\end{table}

\subsection{Reasoning-aware Embedding}
The landscape of IR is rapidly evolving with the emergence of language models capable of complex reasoning and instruction following. Traditional embedding methods, which primarily encode surface-level semantic similarity, often fall short in scenarios requiring logical inference, multi-hop reasoning, or constraint satisfaction. This has motivated the development of \emph{reasoning-aware embedding} techniques—embedding models explicitly designed to encode latent reasoning paths, logical structures, and task-specific constraints into vector representations.

Two complementary research directions highlight the demand for such embeddings: Reasoning-Intensive Retrieval and Instruction-Following IR. Both emphasize retrieval tasks where shallow matching is insufficient, and embeddings must internalize complex relationships.

\textbf{Reasoning-Intensive Retrieval} involves tasks requiring deep inference over queries and documents. Benchmarks such as RAR-b~\cite{xiao2024rar}, BRIGHT~\cite{su2025bright}, and ImpliRet~\cite{taghavi2025impliret} evaluate systems on arithmetic reasoning, implicit fact chaining, and temporal logic. In high-stakes domains like law, datasets like CLERC~\cite{hou2024clerc} and the Legal Reasoning Benchmark~\cite{zheng2025reasoningfocused} demand embeddings that capture legal concepts, precedents, and inferable relationships beyond what surface-level semantics can provide. These challenges necessitate reasoning-aware embedding spaces that can support multi-step deductive or abductive retrieval.

\textbf{Instruction-Following Retrieval}, on the other hand, emphasizes alignment with user intent expressed through natural language instructions. Systems like InstructIR~\cite{oh2024instructira}, MAIR~\cite{sun2024maira}, and InfoSearch~\cite{zhou2025contenta} show that instructions often introduce constraints or goals requiring interpretation and reasoning. Benchmarks such as FollowIR, IFIR~\cite{song2025ifir}, and mFollowIR~\cite{weller2025mfollowir} evaluate how well models can incorporate instruction semantics into retrieval behavior. Embedding-based approaches here must go beyond vanilla query representation—they must fuse instruction semantics and reasoning signals into a unified vector form, again underscoring the need for reasoning-aware embeddings.

These developments collectively reveal a growing shift from simple semantic encoding toward embedding architectures that encode inferential and procedural knowledge, thereby enhancing retrieval in complex, instruction-rich, or high-reasoning domains.








\subsection{Other Domain-Specific Embeddings}
Beyond general-purpose language and multi-modal embeddings, there has been growing interest in developing domain-specific embedding models tailored for specialized fields such as finance and chemistry~\footnote{Code embeddings focus on programming languages and code semantics across tasks, emphasizing data modality rather than a specific application domain, unlike domain-specific embeddings tailored to fields like finance or chemistry.}. These domains pose unique challenges due to their specialized terminology, complex semantics, and domain-specific knowledge requirements. 
The Finance Massive Text Embedding Benchmark (FinMTEB)~\cite{tang2025finmteb} targets the financial domain, covering 64 datasets across seven tasks involving diverse financial text types such as news articles, annual reports, ESG disclosures, and earnings calls in both Chinese and English. FinMTEB evaluations reveal that general-purpose embeddings perform poorly on financial tasks, while domain-adapted models like Fin-E5, trained via persona-based synthetic data, achieve significantly better results. Interestingly, simple Bag-of-Words methods sometimes outperform complex dense embeddings on financial semantic similarity tasks, indicating limitations in current embedding techniques for finance.
Similarly, the Chemical Text Embedding Benchmark (ChemTEB)~\cite{kasmaee2024chemteb} addresses the chemical sciences, a domain with specialized linguistic and semantic challenges. ChemTEB evaluates 34 models on chemical literature and data, providing insights into their ability to handle chemical terminology, molecular descriptions, and scientific language. This benchmark highlights the need for domain-aware embedding strategies in chemistry, as generic models often fall short in capturing domain-specific knowledge. Both benchmarks contribute standardized evaluation frameworks that drive the advancement of more accurate and efficient embeddings tailored for expert domains.

\begin{table*}[ht!]
    \centering
	\caption{Results on MTEB leaderboard as of July 2025. The models are categorized into two groups based on their embedding dimension (Dim.) and the number of parameters (\# Params.). } 
 	\setlength{\tabcolsep}{2pt}
	\resizebox{1.0\linewidth}{!}{
	\begin{tabular}{lcccccccccc}
		\toprule
		& Dim. & \# Params. & Class. & Cluster. & Pair. & Rerank. & Retrieval & STS & Summary & Avg. \\
		\midrule
\multicolumn{11}{l}{Self-Supervised methods} \\   \midrule
		Glove & 200 & - & 57.29 & 27.73 &  70.92 & 43.29 & 21.62 & 61.85 & 28.87 & 41.97 \\
		BERT &  768 & 110M &  61.66 & 30.12 & 56.33 & 43.44 & 10.59 & 54.36 & 29.82 & 38.33\\
		SimCSE-BERT-unsup & 768 & 110M &  62.50 & 29.04 & 70.33 & 46.47 & 20.29 & 74.33 & 31.15 & 45.45 \\ 
		\midrule
		\multicolumn{11}{l}{Supervised methods} \\
		\midrule
		SimCSE-BERT-sup & 768 & 110M &  67.32 & 33.43 & 73.68 & 47.54 & 21.82 & 79.12 & 23.31 & 48.72\\
		Contriever & 768 & 110M &  66.68 & 41.10 & 82.53 & 53.14 & 41.88 & 76.51 & 30.36 & 56.00\\
		SGPT-1.3B & & 1.3B  & 66.52 & 39.92 & 79.58 & 54.00 & 44.49 & 75.74 & 25.44 & 56.11 \\
		GTR-T5-XXL & 768 & 5B & 67.41 & 42.42 & 86.12 & 56.66 & 48.48 & 78.38 & 30.64 & 58.97 \\
		GTR-T5-XL & 768 & 1.2B & 67.11 & 41.51 & 86.13 & 55.97 & 47.96 & 77.80 & 30.21 & 58.42 \\
		Instructor-XL~\cite{su2023one} & 768 & 1.5B & 73.12 & 44.74 & 86.62 & 57.29 & 49.26 & 83.06 & 32.32 & 61.79 \\

		Text-embedding-3-large & 3,072 & n/a & 75.45 & 49.01 & 85.72 & 59.16 & 55.44 & 81.73 & 29.92 & 64.59\\
		E5-mistral-7b-instruct~\cite{wang2023improving} & 4,096 & 7B & 78.47 & 50.26 & {88.34} & 60.21 & 56.89 & {84.63} & {31.40} & 66.63 \\

		{GritLM-7B}~\cite{muennighoff2024generative} & 4,096 & 7B & {79.46} & {50.61} & 87.16 & {60.49} & {57.41} & 83.35 & 30.37 & {66.76} \\

		Echo-mistral-7b-instruct~\cite{springer2024repetition} & 4,096 & 7B & 77.43 & 46.32 & 87.34 & 58.14 & 55.52 & 82.56 & 30.73 & 64.69 \\

		{Gecko}~\cite{lee2024gecko} & 256 & 1.2B & 78.99 & 45.07 & 87.25 & 57.78 & 52.44 & 84.93 & 32.36 & 64.37 \\ 

		LLM2Vec-Llama-3~\cite{behnamghader2024llm2vec} & 4,096 & 8B & 75.92 &46.45 &87.79 & 59.69 & 56.63 &    83.58 & 30.94 &  65.01 \\  
		GTE-Qwen1.5-7B-instruct~\cite{li2023towards} &4,096& 7B & 79.60 & 55.83 & 87.38 &   60.13 & 56.24 &82.42 & 31.46 & 67.34 \\

		SFR-Embedding & 4,096 & 7B &78.33 &51.67 &  88.54 &60.64 &   59.00 & 85.05 & 31.16 & 67.56   \\ 
		Voyage-lite-02-instruct~\cite{voyage2024} &1,024 &-& 79.25 &52.42 & 86.87 & 58.24 & 56.6 &  85.79 & 31.01 & 67.13  \\

		Voyage-large-2-instruct~\cite{voyage2024} &1,024&-& 81.49 & 53.35 &89.24 &60.09 &  58.28 &   84.58 & 30.84 & 68.28 \\
		Linq-Embed-Mistral~\cite{kim2024linq} & 4096 & 7B & 80.2 & 51.4 & 88.4  & 60.3 & 60.2  &  85.0 & 31.0 & 68.2 \\

		NV-Embed~\cite{lee2024nv} & 4,096& 7B & 87.35 & 52.80 & 86.91 &  60.59  & {59.36}  &  82.84 & 31.20 &  {69.32}  \\
		BGE-en-icl~\cite{li2024making} & 4,096 & 7B &  88.95 & 57.89&88.14&59.86&62.16&84.24&30.77 &71.67\\

		NV-Retriever~\cite{moreira2024nv} & 4,096& 7B & 90.37 & 58.46& 88.67 &  60.65  & {62.65}  &  84.31& 30.70 &  {72.31}  \\

        Gemini-embedding~\cite{lee2025gemini}& 3072 & - & 90.05 & 59.39  & 87.7 & 48.59 & 64.35 & 85.29 & 38.28 & 73.3\\

        Qwen-embedding-8B~\cite{zhang2025qwen3} & 1024 & 8B & 90.43 & 58.57 & 87.52 & 51.56 & 69.44 & 88.58 & 34.83 & 75.22 \\
        Qwen-embedding-4B~\cite{zhang2025qwen3} & 2560 & 4B & 89.84 & 57.51 & 87.01 & 50.76 & 68.46 & 88.72 & 34.39 & 74.60 \\
        Qwen-embedding-0.6B~\cite{zhang2025qwen3} & 4096 & 0.6B & 85.76 & 54.05 & 84.37 &48.18 & 61.83 & 86.57 & 33.43 & 70.70\\

	    \bottomrule
	\end{tabular}
	}
 \label{tab:all}
\end{table*}

\section{Discussions}

In this section, we will discuss the performance of representative embedding models on MTEB benchmarks, compare dense and sparse embeddings, and different pooling strategies, and review the implications of scaling laws for larger LLMs.

\vspace{1mm}
\textbf{Performance and Efficiency Comparison.}
Table \ref{tab:all} summarizes the performance of both traditional pretrained models and recent LLM-based models on the MTEB benchmarks. It is evident that supervised methods generally outperform traditional self-supervised methods. Moreover, more recent LLM-based embedding models, such as those built on LLama and Mistral, demonstrate significantly better performance compared to earlier smaller-scale pretrained models like BERT, SGPT, and GTR. This performance boost can be attributed to the increased model size and higher embedding dimensionality of the newer models, which enable them to capture more nuanced semantic relationships and represent richer contextual information within the embeddings. Additionally, through distillation from the output signals of a cross-encoder, a relatively small-scale model like Gecko (1.5B) can achieve performance comparable to that of larger models (7B).
For specific task types, recent LLM-based embedding models demonstrate close performance in pair, reranking, STS, and summary tasks. However, these models (e.g., BGE-en-icl, NV-Embed, NV-Retriever) show significant improvements in clustering, classification, and retrieval tasks, likely due to training datasets that align closely with the MTEB data sources (as seen in Table~\ref{tab:data}). 

\vspace{1mm}
\textbf{Dense vs. Sparse Embedding.}
While constructing dense embeddings based on LLMs has gained significant attention, some researchers are also investigating the acquisition of sparse embeddings by leveraging lexicon-importance distributions derived from LLMs \cite{chen2024bge, zhuang2024promptreps, doshi2024mistral,nie2024text}. \cite{zhuang2024promptreps} evaluate various embedding types by directly prompting or fine-tuning LLMs. Their findings reveal that in zero-shot retrieval settings, dense and sparse embeddings perform differently across different benchmarks: for instance, dense embeddings outperform sparse ones on BEIR and TREC-DL 2019, while sparse embeddings show better results on TREC-DL 2020 and MSMARCO.
However, in supervised fine-tuning settings, dense retrieval significantly outperforms sparse retrieval on datasets like MSMARCO and TREC-DL 2019/2020 when using models such as Llama3-8B and Llama3-70B. Notably, combining both representation types further enhances performance.
\cite{chen2024bge} fine-tune multilingual, multi-granularity text embeddings, integrating both dense and sparse retrieval models based on XLM-RoBERTa. Their results show that dense retrieval notably outperforms sparse retrieval on cross-lingual tasks such as MKQA and MIRACL, and that combining both methods also yields superior results, consistent with the findings reported in \cite{zhuang2024promptreps}.

\vspace{1mm}
\textbf{Last token vs. Mean pooling.}
Two common methods for obtaining embeddings from a sequence of tokens are: i) mean pooling and ii) the embedding of the last [EOS] token. The former calculates the average of token embeddings, which can potentially obscure key information from  phrases. In contrast, [EOS] token embedding is susceptible to recency bias, as it heavily relies on the embedding of the final token in the sequence.
\cite{lee2024nv} conducted experiments comparing several methods for obtaining embeddings from the last layer of LLMs, including the [EOS] token embedding, mean pooling, and their proposed latent attention method. The results demonstrate that mean pooling can always better than [EOS] token embedding based on average MTEB scores, regardless of whether causal attention or bidirectional attention settings are used in the LLMs. Furthermore, their latent attention method enhances embedding capabilities, yielding better results than both the [EOS] token embedding and mean pooling.  This research highlights the potential for exploring more advanced designs to improve last-layer embeddings for enhanced performance.

\vspace{1mm}
\textbf{In-context Learning (ICL) for Embedding Model.} ICL has emerged as a powerful paradigm for enhancing the performance of LLMs in various generative tasks. It enables LLMs to adapt dynamically to the input context by leveraging real-time examples and instructions during inference. This capability to learn from the surrounding context potentially allows LLMs to produce more accurate and context-sensitive embeddings.
Recently, researchers have also explored the effectiveness of ICL for LLM-based representation models. For instance, \cite{jiang2023scaling} first investigated incorporating ICL examples into prompt-based methods and demonstrated their effectiveness for STS tasks. Their approach involved constructing ICL examples by generating word-sentence pairs using ChatGPT to align words with sentence semantics from the STS-B training set, as well as utilizing word-definition pairs from the Oxford dictionary, inspired by DefSent~\cite{tsukagoshi2021defsent}. More recently, \cite{li2024making} incorporated few-shot examples on the query side to enhance query embeddings during supervised contrastive training. Despite these few-shot examples being randomly sampled, they still improved the model's performance.
Besides, \cite{springer2024repetition} proposed repeating the input twice within the context to obtain sequence embeddings, a technique that can be viewed as a specialized form of ICL, and validated its effectiveness in both training-free and supervised contrastive training settings. 
These studies collectively indicate that ICL is a promising strategy for enhancing representation learning with LLMs. Nevertheless, exploring more diverse and high-quality contextual information holds potential for further improving the effectiveness of this approach.

\begin{figure}[t!]
\centering
\includegraphics[width=0.55\linewidth]{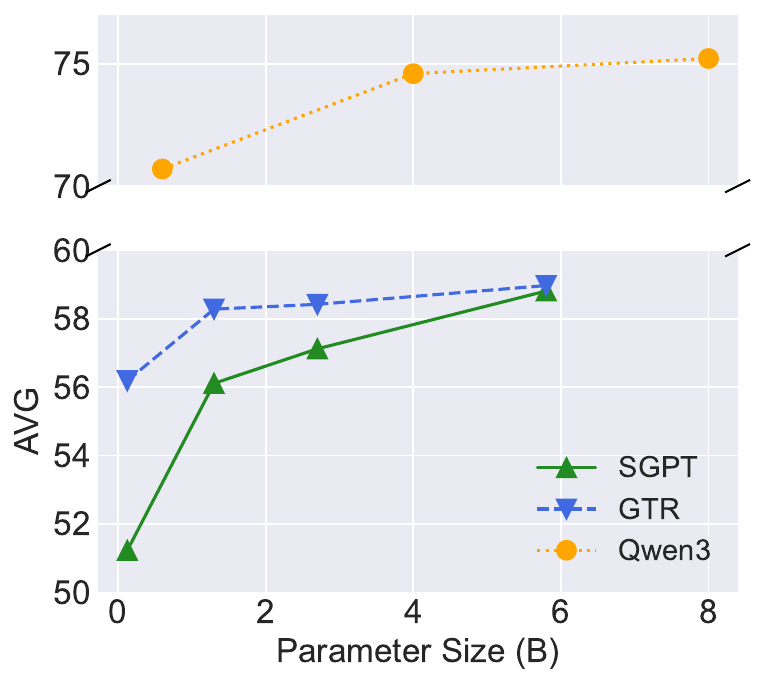}

\caption{The performance trend of models with varying sizes on MTEB benchmark.}
\label{fig:trend}
\end{figure}

\vspace{1mm}
\textbf{Scaling Law of Embedding Models.} 
The question of whether scaling laws hold for embedding models based on LLMS is both significant and intriguing. The effectiveness of embedding models is highly sensitive to several training factors, including model size, data size, and data quality (e.g., diversity and quality of negatives), making it difficult to isolate the effect of each factor independently.
\cite{fang2024scaling} evaluate the quality of dense retrieval models using a contrastive entropy metric and observe a power law relationship between model performance, model size, and data size across various annotation methods and datasets. However, their experiments were conducted on relatively small-scale models, such as BERT models ranging from 0.5M to 110M parameters. 
Additionally, \cite{muennighoff2022mteb} report the performance of models with varying sizes, such as SGPT (125M, 1.3B, 2.7B, 5.8B) and GTR (Base, Large, XL, XXL), on the MTEB benchmark shown in Figure~\ref{fig:trend}, showing consistent improvements as model size increases, even up to 5.8B parameters. \cite{zhuang2024promptreps} further test performance by directly prompting LLMs for embeddings, finding that LLaMA3-70B-Instruct outperforms LLaMA3-8B-Instruct across different embedding types.
Similarly, the Qwen3-Embedding~\cite{zhang2025qwen3} models demonstrate a positive correlation between scale and performance, improving steadily from 0.6B to 4B and 8B; however, the marginal gain from 4B to 8B is notably smaller, suggesting diminishing returns at higher scales.
Taken together, this evidence suggests that the scaling law may also apply to embedding models built on large LMs, although more systematic experiments are needed to rigorously validate this trend across broader configurations.

\section{Open Problems}
Embedding techniques based on LLMs have emerged as a crucial area of research. Despite their impressive performance across various tasks, the process of generating more effective embeddings still presents a number of challenges.
 
\vspace{1mm}
\textbf{Embedding Quality Across Different Tasks.} While LLMs have shown impressive performance in generating embeddings for specific tasks, such as semantic similarity or text classification, their effectiveness across a broader range of tasks remains uncertain. Tasks like clustering, reranking, and summary require embeddings to capture different kinds of relationships between data points, and it is not always clear if embeddings generated by LLMs can consistently perform well in these varied contexts. Moreover, different tasks may prioritize different aspects of the embedding space, such as local versus global structure or task-specific nuances~\cite{cai2022hyper}, which might not be fully captured by general-purpose LLM embeddings. This raises the open question of how LLM embeddings can be optimized or adapted to maintain high-quality representations across diverse tasks~\cite{li2024improvinga}, and whether task-specific tuning or hybrid approaches might be necessary to improve overall effectiveness in more complex and specialized applications.

\vspace{1mm}
\textbf{Efficiency vs. Accuracy Trade-offs.} While LLM-generated embeddings can achieve better accuracy in capturing semantic relationships, they often come at a significant computational cost due to the large model sizes and high-dimensional representations (e.g., 4096 for most recent LLMs). This creates a critical challenge in balancing efficiency and accuracy. For tasks that require real-time processing or deployment in resource-constrained environments, the time and memory demands of using LLMs can be prohibitive. Reducing the computational burden without sacrificing the quality of embeddings is an open problem. Techniques such as model distillation~\cite{lee2024gecko}, pruning, or dimensionality reduction have been proposed, but each comes with trade-offs in terms of performance loss or reduced representational power. Moreover, it remains unclear to what extent the high-dimensional nature of LLM embeddings contributes to their effectiveness, and whether lower-dimensional embeddings could perform comparably in specific tasks. Thus, research is needed to explore strategies that make LLM embeddings more efficient while preserving their robustness, especially in applications where both accuracy and real-time performance are critical. 
Additionally, exploring the embedding performance of relatively smaller models, such as Phi3 \cite{Abdin2024Phi3TR} and MiniCPM \cite{Hu2024MiniCPMUT}, could offer valuable insights into achieving a balance between computational efficiency and embedding quality.

\vspace{1mm}
\textbf{Long Context Embedding.} LLMs are known for their ability to handle long-context dependencies in text, but how well these long contexts are represented in the embeddings they generate is still an open question. Tasks such as document retrieval and multi-hop question answering require embeddings that capture the relationships and themes spanning lengthy passages, yet traditional embedding approaches may struggle to encode this information effectively. One of the main challenges is that long-context embeddings need to balance local context (word-level relationships) with global coherence (document-wide meaning), which is difficult to achieve in a single embedding space. 
Additionally, handling long contexts often increases the computational complexity and dimensionality of the embeddings, making them harder to use in resource-constrained applications. While techniques such as efficient attention mechanisms~\cite{chen2023longlora} and chunk-based processing~\cite{shao2024extensible} have been proposed to address these issues, they often introduce trade-offs by either sacrificing fine-grained contextual information or losing coherence across chunks, limiting their ability to fully capture the intricacies of long-context dependencies
This raises the question of how to design more efficient embedding methods that can faithfully represent long-context information while remaining computationally feasible.

\vspace{1mm}
\textbf{Reasoning Gaps in Embedding-Based Retrieval.}
Despite progress in retrieval modeling, current systems still struggle to perform robustly on reasoning-intensive tasks. A fundamental issue is the lack of reasoning capability in the first-stage retrievers, which are responsible for high-recall candidate generation. These retrievers, typically based on dense or hybrid embeddings, often fail to capture implicit connections, logical dependencies, or multi-step inference chains required for identifying relevant documents. As a result, important evidence is missed early on, limiting the performance ceiling of downstream rerankers or generators. This challenge is especially prominent in benchmarks like BRIGHT and ImpliRet, where retrieval requires understanding implicit facts or combining dispersed clues. The absence of fine-grained supervision for reasoning types during training further compounds the issue, as retrievers are usually optimized with weak relevance signals that do not reflect the nature of the reasoning required. Moreover, standard IR metrics such as recall or nDCG are not designed to evaluate reasoning quality, making progress difficult to measure. Addressing these challenges will require retrieval models that are not only semantically aware but also reasoning-aware—capable of aligning with the latent inference patterns embedded in user queries.

\vspace{1mm}
\textbf{Impact of Training Data Bias on Embeddings.}
LLMs are trained on large text corpora, which often contain subtle biases related to factors like gender, race, and culture~\cite{sakib2024challenging, duan2024large, duan2023ranking}. These biases can inadvertently be reflected in the embeddings generated by LLMs, potentially leading to unfair or unbalanced results in downstream tasks such as sentiment analysis, search, recommendation systems, or hiring platforms. For example, if the training data is skewed towards certain demographics, the embeddings may fail to represent minority groups accurately, resulting in biased outputs. Addressing this issue is challenging, as efforts to reduce bias, such as using de-biasing techniques or filtering the training data, can sometimes come at the cost of model performance or generalization. Furthermore, defining what constitutes “fair” or “unbiased” embeddings is complex and may vary across different tasks and contexts. As a result, balancing fairness with performance remains an open problem. 
Research into better methods for detecting and mitigating bias in LLM-generated embeddings, while preserving their effectiveness in real-world applications, is essential for ensuring that LLMs are more equitable and reliable in diverse use cases.

\vspace{1mm}
\textbf{Adapting LLM Embeddings for Low-Resource Domains.} One of the major challenges in using LLM-based embeddings is their adaptability to low-resource domains, where training data is sparse, or under-representative of the target domain. While LLMs have been pre-trained on large, diverse datasets, these models often struggle when applied to specialized domains, such as medical, legal, or technical fields where pair data is limited~\cite{li2024automir,sivarajkumar2024clinical}. The embeddings generated in these settings may fail to capture the necessary domain-specific nuances, leading to limited performance on downstream tasks like question answering or retrieval. Fine-tuning LLMs for such low-resource domains requires domain-specific data that may not always be available. Additionally, overfitting to small datasets is a concern, which raises the question of how to adapt LLM embeddings to low-resource settings without sacrificing generalization.

\vspace{1mm}
\textbf{Robustness to Adversarial Attacks.} LLM-generated embeddings, while powerful, are often vulnerable to adversarial attacks, where small, carefully crafted perturbations in the input can lead to significant changes in the embeddings and downstream task performance~\cite{chua2024ai,ayyamperumal2024current}. This raises concerns about the robustness of LLM embeddings in real-world applications, where inputs may be noisy or manipulated.  These vulnerabilities pose significant risks, especially in critical applications like healthcare or security information retrieval or retrieval-augmented generation. While there have been some efforts to improve the robustness of embeddings through techniques such as adversarial training, regularization, or perturbation-based testing, these methods are still far from fully addressing the problem. Moreover, ensuring that embeddings remain robust without sacrificing their generalization ability or making them too rigid to handle legitimate variations in input is a complex challenge. The development of more resilient embedding methods that can withstand adversarial manipulations while maintaining high performance across a variety of tasks is an open problem and a vital area for future research.

\vspace{1mm}
\textbf{How to effective training for embedding models?}
LLMs are typically pre-trained on large text corpora for general language modeling tasks, but adapting them specifically for generating high-quality embeddings often requires resource-intensive contrastive training. 
Recent models increasingly rely on large-scale paired data for contrastive fine-tuning; for instance, Qwen3-Emb is trained on 150 million paired examples~\cite{zhang2025qwen3}. 
This poses a significant challenge, particularly in scenarios where only limited data is available. An intriguing open problem is how to effectively tune LLMs for embedding tasks using small-scale datasets. For example, LIMA demonstrates that only about 1,000 high-quality, human-curated samples are sufficient for aligning LLMs in general supervised fine-tuning~\cite{zhou2023lima}. This suggests that efficient fine-tuning with a smaller amount of data might be feasible for embedding tasks as well. Additionally, exploring whether there are more effective training methods or strategies tailored specifically for tuning LLMs as embedding models could provide valuable insights.

\vspace{1mm}
\textbf{Is Continued Training Beneficial for Embedding Models?}
Current LLMs are pretrained primarily on next-token prediction (NTP) tasks, which focus on sequence completion rather than embedding generation. This misalignment raises the question of whether continued training with alternative tasks, such as next sentence prediction or sentence-level reconstruction, could enhance their performance as embedding models.  Future research could explore how and what domain-specific or task-specific continued training objectives can be designed to preserve generalization while improving domain relevance, balancing computational efficiency, embedding quality, and downstream task adaptability.

\section{Conclusion}
In conclusion, this survey addresses the paradigm shift toward leveraging LLMs as embedding models, emphasizing their impact on representation learning across diverse tasks in natural language understanding, information retrieval and recommendation. The paper provides an in-depth analysis of both tuning-free and tuning-based approaches for generating effective embeddings, highlighting methods that maximize LLM capabilities without extensive fine-tuning, as well as techniques that further finetune these models for task-specific contexts. By comparing various LLM-based embedding strategies, including dense and sparse embeddings, pooling techniques, and scalability considerations, we provide practical insights for researchers and practitioners. Moreover, this survey identifies key open challenges. As LLMs continue to evolve, this work aims to serve as a valuable guide for understanding and advancing embedding methodologies in the LLM era, ultimately supporting improved performance across a wide range of NLP and IR applications.



\begin{acks}
We thank the editors and the anonymous reviewers for their constructive comments and advice. 
\end{acks}

\bibliographystyle{ACM-Reference-Format}
\bibliography{acmart,custom}










\end{document}